\documentclass[10pt,twocolumn,letterpaper]{article}

\usepackage{wacv}
\usepackage{times}
\usepackage{epsfig}
\usepackage{graphicx}
\usepackage{amsmath}
\usepackage{amssymb}
\usepackage{url}
\usepackage{booktabs}
\usepackage{times}
\usepackage{multirow}
\usepackage{array}
\usepackage{algorithm}
\usepackage{algpseudocode}
\usepackage{subcaption}
\usepackage{stmaryrd}
\usepackage{xcolor, colortbl}
\usepackage[font=small,labelfont=bf]{caption}
\usepackage{pifont} 
\usepackage{xr}
\newcommand{\cmark}{\ding{51}}%
\newcommand{\xmark}{\ding{55}}%

\def\wacvPaperID{672} 

\wacvfinalcopy 

\ifwacvfinal
\fi

\usepackage{cite}

\ifwacvfinal
\usepackage[breaklinks=true,bookmarks=false]{hyperref}
\else
\usepackage[pagebackref=true,breaklinks=true,colorlinks,bookmarks=false]{hyperref}
\fi

\newcommand{\Eqref}[1]{Eq.~(\ref{#1})}
\newcommand{\ours}{S${}^3$D\xspace}
\definecolor{grey}{rgb}{0.9, 0.9, 0.9}
\newcommand{\ccol}{\cellcolor{grey}}

\begin{document}

\title{Semi-supervised Domain Adaptation via Sample-to-Sample Self-Distillation}

\author{
Jeongbeen Yoon
\and
Dahyun Kang
\and
Minsu Cho
\vspace{+0.1mm}
\and
\vspace{+1mm}
Pohang University of Science and Technology (POSTECH), South Korea
\\
{\tt\small \{jeongbeen,dahyun.kang,mscho\}@postech.ac.kr}
}

\maketitle
\thispagestyle{empty}

\ifwacvfinal
\pagestyle{empty}
\fi


\begin{abstract}
Semi-supervised domain adaptation (SSDA) is to adapt a learner to a new domain with only a small set of labeled samples when a large labeled dataset is given on a source domain.
In this paper, we propose a pair-based SSDA method that adapts a model to the target domain using self-distillation with sample pairs. 
Each sample pair is composed of a teacher sample from a labeled dataset (\ie, source or labeled target) and its student sample from an unlabeled dataset (\ie, unlabeled target). Our method generates an assistant feature by transferring an intermediate style between the teacher and the student, and then train the model by minimizing the output discrepancy between the student and the assistant.
During training, the assistants gradually bridge the discrepancy between the two domains, thus allowing the student to easily learn from the teacher.
Experimental evaluation on standard benchmarks shows that our method effectively minimizes both the inter-domain and intra-domain discrepancies, thus achieving significant improvements over recent methods.
\end{abstract}

\vspace{-3mm}
\section{Introduction}
\begin{figure}
\centering
\includegraphics[width=0.48\textwidth]{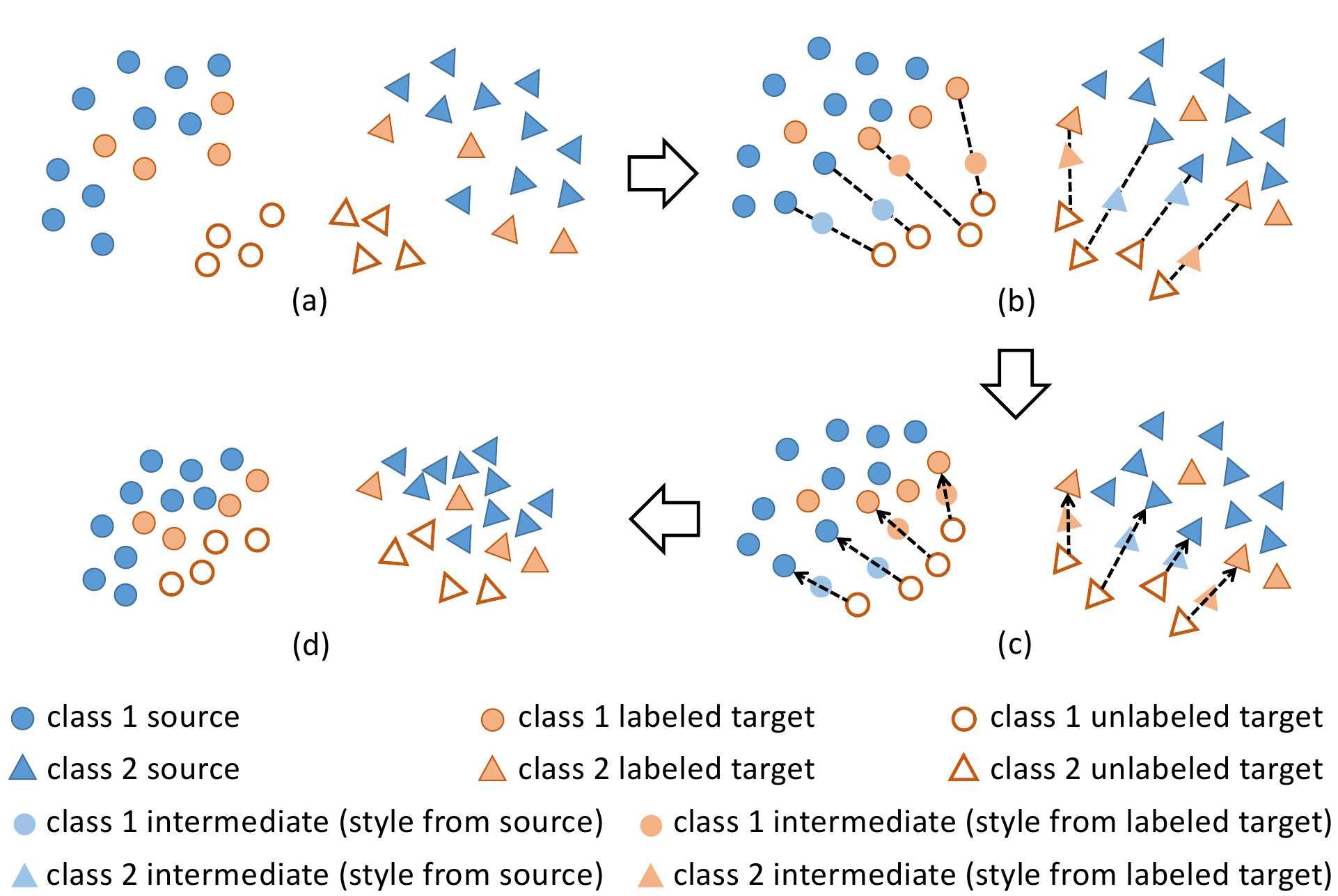}
\vspace{-4mm}
\caption{
Sample-to-sample self-distillation (\ours). 
(a) Before adaptation, both inter-domain and intra-domain discrepancy are present. 
(b) \ours composes sample pairs of a teacher (labeled one) and a student (unlabeled one), and generates assistants by transferring an intermediate style of each pair. 
(c) The assistants bridge the discrepancy between the two domains, thus facilitating domain adaptation by self-distillation.
(d) As the result, \ours reduces both the inter- and intra-domain discrepancy. See Section~\ref{sec:sssd} for detail. 
}
\label{fig:teaser}
\vspace{-0.5cm}
\end{figure}
Deep neural networks have shown impressive performance in learning tasks on a domain where a large number of labeled data are available for training \cite{krizhevsky2012imagenet, simonyan2014very, he2016deep}. 
However, they often fail to generalize to a new domain where the distribution of input data significantly deviates from the original domain, \ie, when a domain gap arises.    
The goal of domain adaptation is to adapt a learner to the new domain ({\em target}) using the labeled data available from the original domain ({\em source}).
Unsupervised domain adaptation (UDA) attempts to tackle this inter-domain discrepancy problem without any supervision on the target domain, assuming that no labels for samples are available from the target domain in training \cite{ganin2016domain, saito2017adversarial, long2018conditional, hoffman2018cycada}.
In contrast, semi-supervised domain adaptation (SSDA) relaxes the strict constraint, using a small number of additional labels on the target data, \eg, a few labels per class \cite{saito2019semi}. As we are able to obtain such additional labels easily on the target data, it renders the adaptation problem more practical and better situated in learning.

\begin{figure}
\centering
\includegraphics[width=0.3\textwidth]{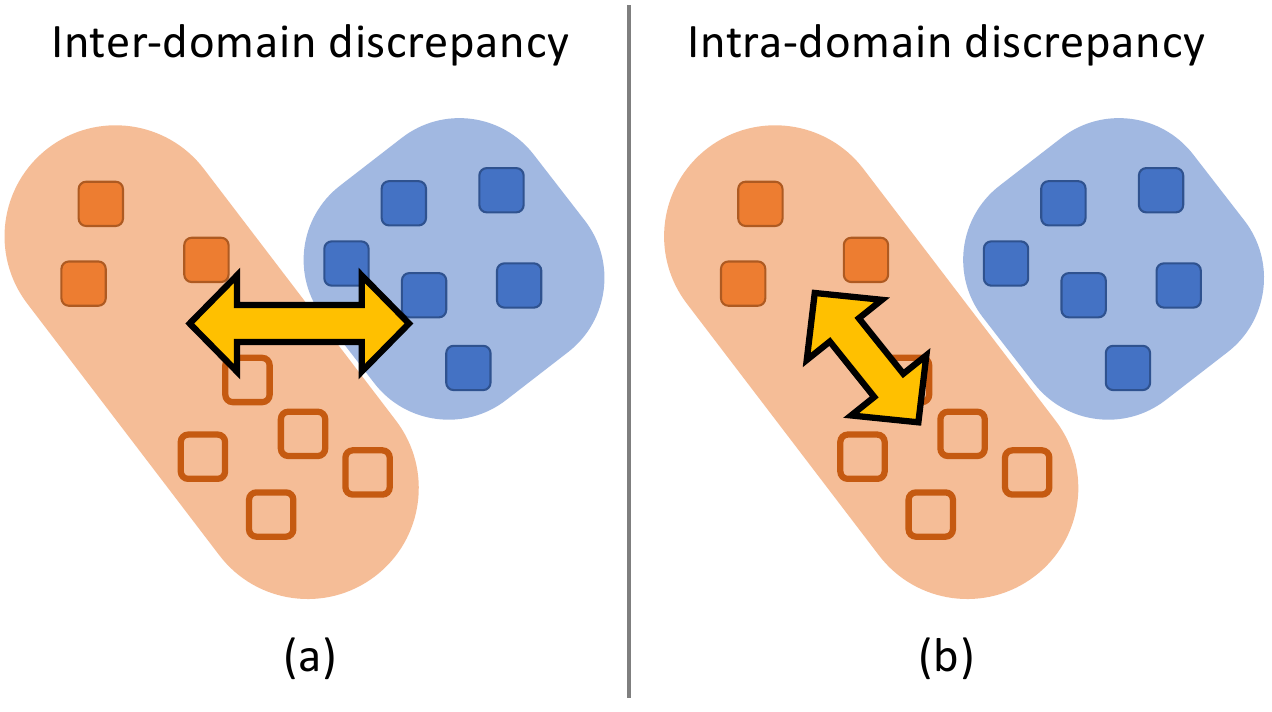}
\caption{
Two problems of SSDA.
An Orange region represents a target domain and a blue region represents a source domain.
} 
\label{fig:embedding}
\vspace{-0.5cm}
\end{figure}

Empirical results \cite{saito2019semi} show that a na\"ive adaptation of UDA to SSDA, \eg, considering the labeled samples on the target domain as a part of those on the source domain, suffers from the effect of target intra-domain discrepancy, \ie, the distribution of labeled samples on the target domain is separated from that of unlabeled samples during training.
We consider the intra-domain discrepancy and the aforementioned inter-domain discrepancy as major challenges of SSDA, and we illustrate them in Figure~\ref{fig:embedding}.
Previous methods for SSDA \cite{saito2019semi, kim2020attract} aims to address the issue using a prototype-based approach; they create a prototype representation for each class and reduce the distance between each prototype and its nearby unlabeled samples. 

In this paper, we propose a new SSDA approach, dubbed {\em sample-to-sample self-distillation} (\ours), that leverages rich sample-to-sample relations rather than prototype-to-data relations.
Our method takes a labeled sample as a teacher on either source or target domain and an unlabeled sample as a student.
When the teacher comes from the source domain, it minimizes the inter-domain discrepancy between the source and the target. 
When the teacher is a labeled sample on the target domain, it effectively suppresses the intra-domain discrepancy within the target. 
For the reason that na\"ively reducing the domain gap is demanding, we generate assistant features which support to bridge the domain gap.
It is known that the domain and the style of an image are closely related and training with mixed style features helps to bridge the domain discrepancy \cite{zhou2021domain, gong2019dlow}.
Inspired by the fact, the assistant features are created by transferring intermediate styles between the teacher and the student.
Then a model is trained by minimizing the output discrepancy between the assistant and the student.
The assistant features smoothly bridge the discrepancy between the two domains, thus making the student easily learns from the teacher.

To generate reliable pairs of the teacher and the student, we employ pseudo-labeling~\cite{lee2013pseudo} and present a new form of reliability evaluation on the pseudo-label motivated by \cite{zhang2019category}.
Compared to the previous prototype-based approach, our pair-based approach fully exploits rich and diverse supervisory signals via data-to-data distillation and effectively adapts to the target domain by minimizing both the intra-domain and inter-domain discrepancy. The contributions of this paper are summarized as follows:
\begin{itemize}
    \item We propose {\em sample-to-sample self-distillation} (\ours) that exploits rich sample-to-sample relations using self-distillation.
    \item We generate assistant features of which the style is represented by an intermediate of the source and the target to fill the domain gap, thus facilitating the adaptation.
    \item We show that \ours effectively adapts a network to a target domain by alleviating both the inter- and intra-domain discrepancy and \ours sets a new state of the art.\looseness=-1
\end{itemize}

\begin{figure*}
\centering
\includegraphics[width=0.72\textwidth]{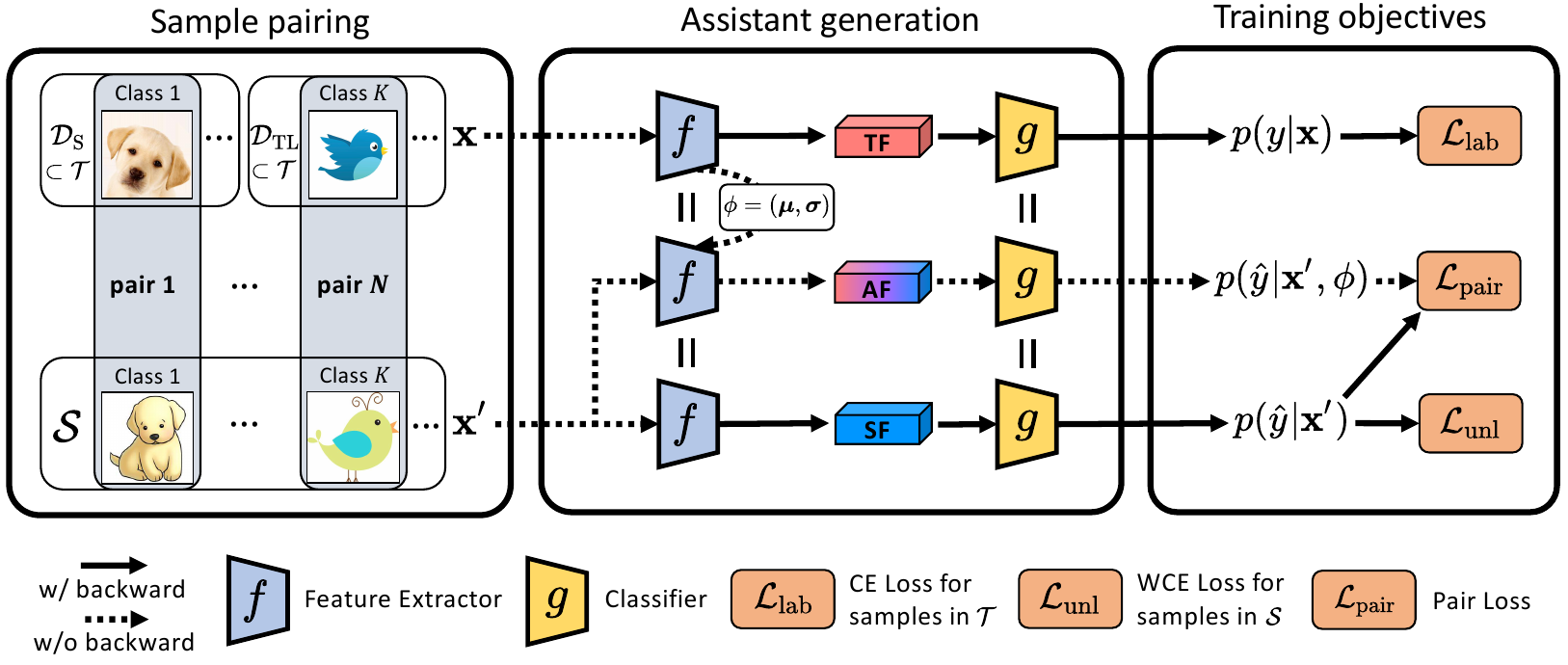}
\vspace{-1mm}
\caption{Overview of the Sample-to-Sample Self-Distillation (\ours). 
We use a feature extractor and a classifier trained in the pre-training stage (Section~\ref{sec:classifier_and_pretraining}). 
\ours consists of the reliable student-set generation step (RSS), and the sample pairing, assistant feature generation and self-distillation step.
We omit the reliable student-set generation step (RSS) in the figure.
The abbreviations of TF, AF, and SF represent the teacher, the assistant, and the student feature, respectively.
In the sample pairing step, the model pairs a student and a teacher of the same (pseudo-)label.
In the assistant generation step, the teacher is forwarded and the set of mean and standard deviation of the teacher is extracted.
The extracted $\phi$ is utilized for adjusting intermediate style to the student to generate the assistant feature.
The $\mathcal{L}_\text{pair}$ is then calculated to minimize the difference between the outputs of the assistant and the student.
We omit feature normalization, temperature scaling, and softmax operation for simplicity.
}\label{fig:architecture}
\vspace{-0.5cm}
\end{figure*}
\vspace{-3mm}
\section{Related work}

\label{related-work}
\noindent
\textbf{Semi-supervised domain adaptation.} 
The goal of semi-supervised domain adaptation (SSDA) is to adapt a model on the target domain with a few labels of target data \cite{saito2019semi}. 
Although SSDA has been considered in \cite{ao2017fast, donahue2013semi, yao2015semi}, most recent research has explored unsupervised domain adaptation (UDA). 
The main issue of domain adaptation is the gap between the source and the target domain distributions. 
Previous UDA methods focus on aligning the two domain distributions. 
Adversarial learning between a domain-classifier and a feature extractor is one of the representative UDA approaches \cite{ganin2016domain, saito2017adversarial, long2018conditional, lee2019drop, xu2019adversarial}. 
Learning with pseudo-labels~\cite{lee2013pseudo} is another approach in UDA~\cite{xie2018learning,chang2019domain,deng2019cluster,zhang2019category}. 
To supplement the absence of target domain labels, the network assigns labels to the target data in a certain standard.
The network then utilizes the obtained pseudo-labels as supervision for training using the target domain data.
SSDA is re-examined in Minimax Entropy (MME)~\cite{saito2019semi} for taking the advantage of extra supervision.
With a minor effort, the model benefits from just a few target labels.
MME discovers the ineffectiveness of previous UDA methods in SSDA, and proposes a new approach for the task. 
They minimize the distance between the class prototypes and nearby unlabeled target samples by minimax entropy.
After MME, several new SSDA methods are followed. 
\cite{jiangbidirectional} generate bidirectional adversarial samples from source to target domain and from target to source domain to fill the domain gap.
Attract, Perturb, and Explore (APE)~\cite{kim2020attract} analyzes the target intra-domain discrepancy issue, and suggests to minimize the gap using Maximum Mean Discrepancy, perturbation loss, and the class prototypes.
Among the previous work, MME and APE use the class prototypes and adapt to the target domain for SSDA.
We tackle the issues of SSDA in a simple pair-based way by applying self-distillation different from previous work.
Dissimilar to the prototype-based way, our pair-based method enables unlabeled target samples to be trained with more abundant supervision.

\noindent
\textbf{Style Manipulation.}
The style of images has been manipulated to increase the recognition ability of neural networks~\cite{hoffman2018cycada, gong2019dlow, ulyanov2016instance, huang2017arbitrary, zhou2021domain}.
Previous work of style transfer~\cite{ulyanov2016instance, huang2017arbitrary} find that mean and standard deviation of an intermediate feature from neural networks are closely related to the style of an image.
Further, \cite{zhou2021domain} reveals that domain is related to the image style and suggest to mix the style of given source domains to generalize the model in domain generalization.
Our method is motivated by the fact that the intermediate domain style helps to minimize the domain discrepancy~\cite{gong2019dlow, zhou2021domain}.
Unlike \cite{zhou2021domain}, which trains the model by exposing it to various styles from source domains, our method generates intermediate style features (assistants) from labeled (teachers) and unlabeled samples (students) to guide students.
Also, we use the assistant only for matching its soft output with the student's one.
Therefore, our method forces to produce the same results between two features of the same content with different styles.

\noindent
\textbf{Knowledge distillation.} 
The idea of knowledge distillation (KD) is to train a model \emph{(student)} by transferring knowledge extracted from another model \emph{(teacher)} that is more powerful than the student~\cite{breiman1996born, bucilua2006model}.
A series of study on KD has shown its attractive characteristics such as regularizing the student~\cite{yuan2020revisiting}, stabilizing training~\cite{cheng2020explaining}, and preventing models to be overconfident~\cite{yun2019regularizing}.
One line of work on KD assumes two independent teacher and student networks sharing an input sample, and maps the output of the student to that of the teacher~\cite{hinton2015distilling, romero2014fitnets, zagoruyko2016paying, park2019relational}.
This branch of work motivates GDSDA~\cite{ao2017fast}, which proposes to use multiple pre-trained source models to give predictions to a target model for domain adaptation tasks.
The other interesting line of work on KD investigates self-knowledge distillation; a single network is trained by the knowledge from itself~\cite{furlanello2018born, xie2020self, yun2019regularizing}.
Our design resorts to the second line of work.
We propose to minimize Kullback–Leibler divergence of two predictions between an intermediate style feature and its corresponding unlabeled target sample in the form of self-distillation.
This learning objective naturally conforms to the goal of domain adaptation: adapting a learner to a target domain by aligning two samples sharing semantics yet visually diverse.


\section{Method}
\label{methods}

The task of semi-supervised domain adaptation is formulated as to classify unlabeled samples on a target domain using labeled samples on a source domain together with  a limited number of labeled samples on the target domain~ \cite{saito2019semi, kim2020attract, li2020online}. 
Let us consider three datasets given in this context: a source dataset $\mathcal{D}_{\text{S}} = \{(\mathbf{x}_\text{S}^{(i)}, y_\text{S}^{(i)})\}_{i=1}^{N_\text{S}}$, a labeled target dataset $\mathcal{D}_{\text{TL}} = \{(\mathbf{x}_\text{TL}^{(j)}, y_\text{TL}^{(j)})\}_{j=1}^{N_\text{TL}}$, and an unlabeled target dataset $\mathcal{D}_{\text{TU}} = \{ \mathbf{x}_\text{TU}^{(k)} \}_{k=1}^{N_\text{TU}}$, 
where $\mathbf{x}$, $y$, and $N$ denote a sample, its corresponding label, and the number of samples, respectively.
Here, we are given only a limited number of labeled samples per class on the target domain, \ie, $N_\text{TL} \ll N_\text{TU}$.
The two domains share the same label space $y \in \{1, \cdots, K\}$ but with different input distribution.
In this setup, we train a model on $\mathcal{D}_{\text{train}} = \mathcal{D}_{\text{S}} \cup \mathcal{D}_{\text{TL}}$, and $\mathcal{D}_{\text{TU}}$, and then evaluate it on $\mathcal{D}_{\text{test}} = \mathcal{D}_{\text{TU}}$ with its ground-truth labels. 
In training, we validate models on additional labeled target set $\mathcal{D}_{\text{val}}$ of $\mathcal{D}_{\text{val}} \cap \mathcal{D}_{\text{train}} = \mathcal{D}_{\text{val}} \cap \mathcal{D}_{\text{test}} = \emptyset$.
We select the best model and search hyper-parameters on the validation set.

\vspace{-1mm}
\subsection{Classifier model and its pre-training}
\label{sec:classifier_and_pretraining}
Our model consists of two parts: a feature extractor $f(\cdot;\boldsymbol \theta)$ and a classifier $g(\cdot;\mathbf{W})$, where $\boldsymbol \theta$ and $\mathbf{W}$ denote trainable parameters. 
We use a convolutional neural network for $f(\cdot;\boldsymbol \theta)$, and a distance-based classifier for $g(\cdot;\mathbf{W})$~\cite{wang2018cosface, chen2019closer}. 
The distance-based classifier computes its output as the cosine similarity between the input feature $\mathbf{h}$ and each column $\mathbf{w}_k$ of $\mathbf{W}$: 
\vspace{-1mm}
\begin{align}
    p(y&|\mathbf{x}) =  \text{softmax}\left(\frac{g(f(\mathbf{x}; \boldsymbol \theta); \mathbf{W})}{T}\right), \nonumber \\ 
& \text{where}\,\,
g(\mathbf{h}; \mathbf{W}) = \left[\frac{\mathbf{w}_1}{\lVert \mathbf{w}_1 \rVert}
;\cdots;\frac{\mathbf{w}_K}{\lVert \mathbf{w}_K \rVert}\right]^{\top}\frac{\mathbf{h}}{\lVert \mathbf{h}\rVert}, 
\end{align}
where the final prediction $p(y|\mathbf{x})$ is obtained via softmax operation with a temperature $T$.
In the following subsections, we often omit the function parameters, $\boldsymbol \theta$ and $\mathbf{W}$, for notational simplicity. 

We pre-train the model with labeled samples in $(\mathbf{x}, y) \in \mathcal{T}$, where $ \mathcal{T} = \mathcal{D}_{\text{S}} \cup \mathcal{D}_{\text{TL}}$, via minimizing the cross-entropy loss:
\vspace{-2mm}
\begin{eqnarray}
\label{eq:ce_loss}
\mathcal{L}_\text{lab} = \mathbb{E}_{(\mathbf{x},y)}[-\log p(y|\mathbf{x})].
\end{eqnarray}
This pre-training improves the performance of \ours and also speeds up its convergence.

\floatname{Algorithm}{Training process of \ours}
\renewcommand{\algorithmicrequire}{\textbf{Input:}}
\renewcommand{\algorithmicensure}{\textbf{Output:}}
\begin{algorithm}[t!]
    \caption{Sample-to-Sample Self-Distillation.}
    \label{SSSD-algorithm}
    \begin{algorithmic}[1] 
        \Require $\mathcal{T}$: Teacher set
        \Require $\boldsymbol \theta$ and $\mathbf{W}$: Pre-trained weights 
        \Require $M$: student-set generation interval
        \For{$e \gets \ 1 \ \text{to} \ \texttt{max}\_\texttt{steps}$}
            \If{$e \bmod M \ \text{is} \ 0$}
                \State Update student set $\mathcal{S}$ \algorithmiccomment{\Eqref{eq:reliability_evaluation}}
            \EndIf
            \State $(\mathbf{x}, y) \ \mathtt{\sim} \ \mathcal{T}$
            \State $(\mathbf{x}', \hat{y}) \ \mathtt{\sim} \ \mathcal{S} \quad \text{such that } \ \hat{y} = y$
            \State $\phi = (\boldsymbol{\mu}, \boldsymbol{\sigma}) \gets \text{get feature statistics from }f(\mathbf{x};\boldsymbol \theta)$
            \State $p(y|\mathbf{x}) \gets \text{softmax}(g(f(\mathbf{x};\boldsymbol \theta);\mathbf{W}) / T)$  
            \State $p(\hat{y}|\mathbf{x}',\phi) \gets \text{softmax}(g(f'(\mathbf{x}';\boldsymbol \theta, \phi);\mathbf{W}) / T)$
            \State $p(\hat{y}|\mathbf{x}') \gets \text{softmax}(g(f(\mathbf{x}';\boldsymbol \theta);\mathbf{W}) / T)$ 
            \State $\mathcal{L}_\text{lab} \gets \text{CE}(p(y|\mathbf{x}), y)$ \algorithmiccomment{\Eqref{eq:ce_loss}.}
            \State $\mathcal{L}_\text{unl} \gets \text{WCE}( p(\hat{y}|\mathbf{x'})$, $\hat{y})$  \algorithmiccomment{\Eqref{eq:wce_loss}.}
            \State $\mathcal{L}_\text{pair} \gets D_{\text{KL}}(p(\hat{y}|\mathbf{x}',\phi)$, $p(\hat{y}|\mathbf{x}'))$  \algorithmiccomment{\Eqref{eq:pair_loss}.}
            \State $\mathcal{L} \gets \mathcal{L}_\text{lab} + \mathcal{L}_\text{unl} + \lambda\mathcal{L}_\text{pair}$ 
            \State update $\boldsymbol \theta$ and $\mathbf{W}$ with $\mathcal{L}$ using SGD
            \If{$e \bmod \ \texttt{val}\_\texttt{freq} \ \text{is} \ 0$}
                \State \text{validate and early-stop} 
            \EndIf
        \EndFor
    \end{algorithmic}
\end{algorithm}

\subsection{Sample-to-sample self-distillation (\ours)}
\label{sec:sssd}
The {\em sample-to-sample self-distillation} (\ours) is designed to perform SSDA by simultaneously minimizing both the inter-domain discrepancy (between the source and the target) and the intra-domain discrepancy (within the target). 
It achieves the goal by alternating two steps: the \emph{student-set generation step}, and the training step.
The training step consists of \emph{sample pairing, assistant generation and self-distillation}.    
At the student-set generation step, we pseudo-label samples from unlabeled target dataset $\mathcal{D}_{\text{TU}}$ and select reliable ones using reliability evaluation. The resultant set $\mathcal{S}$ is used for student samples in self-distillation. 
At the training step, we randomly produce teacher-student pairs with the same class label, generating corresponding assistant features, and perform self-distillation by minimizing the distance between the assistant and the student predictions. In pairing, we take one sample from $\mathcal{T}$ (as a teacher) and the other from $\mathcal{S}$ (as a student). 
The overall procedure is summarized in Algorithm~\ref{SSSD-algorithm} and also illustrated in Figure~\ref{fig:architecture}.
In the following, we explain the details of each step and describe the overall training objective. \\

\vspace{-2mm}
\noindent
\textbf{Reliable student-set generation.}
\label{pseudo-labeling}
This step consists of pseudo-labeling and reliability evaluation. 
We assign a class label $\hat{y}$ to each unlabeled sample $\mathbf{x}' \in \mathcal{D}_{\text{TU}}$, and construct a pseudo-labeled set of the student samples $\{(\mathbf{x}'^{(l)}, \hat{y}^{(l)})\}_{l=1}^{N_\text{TU}}$; we simply take a pseudo-label $\hat{y}$ of $\mathbf{x}'$ as the class index $k$ of the maximum prediction value:
\begin{eqnarray}
\label{eq:pseudo-labeling}
\hat{y} = \operatorname*{argmax}_{k \in \{1, 2, \dots , K\}} p(y=k|\mathbf{x}').
\end{eqnarray}
Although pseudo-labeling enables supervised training on unlabeled samples, pseudo-labels are often incorrect, in particular, in an early stage of training.
We thus drop unreliable samples to compose $\mathcal{S}$ for pairing.
Let $\pi_j(\cdot)$ be a selection operator that selects $j^{\text{th}}$ largest value.
We construct a student set by reliability evaluation:
\begin{align}
\mathcal{S} = \{(\mathbf{x}', \hat{y}) | &(\pi_1(g(f(\mathbf{x'}))) - \pi_2(g(f(\mathbf{x'}))) > \delta) \nonumber\\
&\lor (\pi_1(p(\hat{y}|\mathbf{x}'))> \alpha) ; \forall \mathbf{x}' \in \mathcal{D}_{\text{TU}} \},
\label{eq:reliability_evaluation}
\end{align}
where $\delta$ is an average margin between the first and the second highest logits of all unlabeled target on the pre-trained model, and $\alpha$ is a hyper-parameter.
We include analysis on the margin $\delta$ in the supplementary materials in detail.
The first condition is met when a margin between the largest and the second largest value of the logit is high enough~\cite{zhang2019category}.
The second condition is met when the absolute largest class probability score is high enough.
In this way, the model assigns pseudo-labels to confident samples only and generates reliable student-set (RSS) so that the model can take reliable pairs. 
\newline

\begin{table*}[t!]
\begin{center}
\scalebox{0.85}{
\begin{tabular}{l|l|cccccccccccccc|cc}
\toprule[1pt] 
 \multirow{2}{*}{Net} & \multirow{2}{*}{Method}       &\multicolumn{2}{c}{R to C}&\multicolumn{2}{c}{R to P} & \multicolumn{2}{c}{P to C}  & \multicolumn{2}{c}{C to S} & \multicolumn{2}{c}{S to P} & \multicolumn{2}{c}{R to S} & \multicolumn{2}{c}{P to R}     &\multicolumn{2}{|c}{MEAN} \\ 
& &1\scriptsize{-shot}&3\scriptsize{-shot} &1\scriptsize{-shot}&3\scriptsize{-shot}&1\scriptsize{-shot}&3\scriptsize{-shot} &1\scriptsize{-shot}&3\scriptsize{-shot}&1\scriptsize{-shot}&3\scriptsize{-shot}&1\scriptsize{-shot}&3\scriptsize{-shot} &1\scriptsize{-shot}&3\scriptsize{-shot} &1\scriptsize{-shot}&3\scriptsize{-shot}  \\ \hline
 \multirow{9}{*}{\rotatebox[origin=c]{90}{AlexNet}} 
&S+T& 43.3   & 47.1 & 42.4   & 45.0 & 40.1   & 44.9 & 33.6   & 36.4 & 35.7   & 38.4 & 29.1 & 33.3 & 55.8   & 58.7 & 40.0 & 43.4 \\
&DANN & 43.3   & 46.1 & 41.6   & 43.8 & 39.1   & 41.0 & 35.9   & 36.5 &36.9   & 38.9 & 32.5 & 33.4 & 53.6   & 57.3 & 40.4 & 42.4 \\
&ADR & 43.1 & 46.2 & 41.4  & 44.4 & 39.3  & 43.6  & 32.8  &  36.4   & 33.1  & 38.9   & 29.1   & 32.4 & 55.9  & 57.3 & 39.2  & 42.7 \\
&CDAN & 46.3   & 46.8 & 45.7   & 45.0 & 38.3   & 42.3 & 27.5   & 29.5 & 30.2   & 33.7 & 28.8 & 31.3 & 56.7   & 58.7 & 39.1 & 41.0 \\
&ENT & 37.0   & 45.5 & 35.6   & 42.6 & 26.8   & 40.4 & 18.9   & 31.1 & 15.1   & 29.6 & 18.0 & 29.6 & 52.2   & 60.0 & 29.1 & 39.8 \\
&MME & 48.9 &  55.6 & 48.0   &  49.0 &  46.7   & 51.7 &  36.3   & 39.4 &  39.4  & 43.0 & 33.3 & 37.9 & 56.8  & 60.7 & 44.2 & 48.2\\
&APE & 47.7 & 54.6 & 49.0 & 50.5 & 46.9 & 52.1 & 38.5 & 42.6 & 38.5 & 42.2 & 33.8 & 38.7 & 57.5 & 61.4 & 44.6 & 48.9\\\cline{2-18}
& \ours w/o AF & 52.3 & 56.2 & 48.7 & 51.2 &  48.0 & 51.3 &  39.2 & 43.5 &  40.6 &  46.5 &  37.4 &  \bf{39.8} &  59.5 &  65.1 & 46.5 &  50.5  \\
& \ccol\ours (ours)  &\ccol \bf{53.5} &\ccol \bf{56.5} &\ccol \bf{51.8} &\ccol \bf{52.2} &\ccol \bf{49.1} &\ccol \bf{53.9} &\ccol \bf{40.1} &\ccol \bf{44.4}  &\ccol \bf{44.9} &\ccol \bf{48.7} &\ccol \bf{39.9} &\ccol 39.2 &\ccol \bf{61.7} &\ccol \bf{65.4} &\ccol \bf{48.7} &\ccol \bf{51.5} \\\hline\hline
\multirow{9}{*}{\rotatebox[origin=c]{90}{ResNet34}} 
&S+T    & 55.6 & 60.0   & 60.6 & 62.2   & 56.8 & 59.4   & 50.8 & 55.0   & 56.0 & 59.5 & 46.3 & 50.1   & 71.8 & 73.9 & 56.9 & 60.0 \\
&DANN & 58.2 & 59.8   & 61.4 & 62.8   & 56.3 & 59.6   & 52.8 & 55.4   & 57.4 & 59.9 & 52.2 & 54.9   & 70.3 & 72.2 & 58.4 & 60.7 \\
&ADR & 57.1 & 60.7 & 61.3 & 61.9 & 57.0 & 60.7 & 51.0 & 54.4 & 56.0 & 59.9 & 49.0 & 51.1 & 72.0 & 74.2 & 57.6 & 60.4 \\
&CDAN & 65.0 & 69.0   & 64.9 & 67.3   & 63.7 & 68.4   & 53.1 & 57.8   & 63.4 & 65.3 & 54.5 & 59.0   & 73.2 & 78.5 & 62.5 & 66.5 \\
&ENT & 65.2 & 71.0   & 65.9 & 69.2   & 65.4 & 71.1   & 54.6 & 60.0   & 59.7 & 62.1 & 52.1 & 61.1   & 75.0 & 78.6 & 62.6 & 67.6 \\
&MME & 70.0 & 72.2   & 67.7 & 69.7   & 69.0 & 71.7   & 56.3 & 61.8   & 64.8 & 66.8 & 61.0 & 61.9   & 76.1 & 78.5 & 66.4 & 68.9\\
&APE & 70.4 & 76.6 & \bf{70.8} & \bf{72.1} & 72.9 & 76.7 & 56.7 & 63.1 & 64.5 & 66.1 & 63.0 & 67.8 & 76.6 & 79.4 & 67.6 & 71.7 \\\cline{2-18}
& \ours w/o AF  & \bf{73.4} & 75.3 &   69.2 &   70.8 &   \bf{73.4} &   74.4 &   60.2 &   63.1 &   66.1 &   69.1 &   62.8 & 64.7 &   79.3 & 79.7 & 69.2 & 71.0 \\
& \ccol \ours (ours)  &\ccol 73.3 &\ccol \bf{75.9} &\ccol 68.9 &\ccol \bf{72.1} &\ccol \bf{73.4} &\ccol \bf{75.1} &\ccol \bf{60.8} &\ccol \bf{64.4} &\ccol \bf{68.2} &\ccol \bf{70.0} &\ccol \bf{65.1} &\ccol \bf{66.7} &\ccol \bf{79.5} &\ccol \bf{80.3} &\ccol \bf{69.9} &\ccol \bf{72.1} \\
\bottomrule[1pt]
    \end{tabular}}
\end{center}
\vspace{-4mm}
\caption{Classification accuracy on the DomainNet dataset (\%) for one-shot and three-shot on four domains (R: Real, C: Clipart, P: Painting, S: Sketch). $\dagger$ denotes that we reproduce the baseline. \label{tb:domain_all}}
\vspace{-3mm}
\end{table*}

\noindent
\textbf{Sample pairing and assistant generation.}
After obtaining $\mathcal{S}$, we construct a pair of a labeled sample $(\mathbf{x}, y) \in \mathcal{T}$ and a pseudo-labeled sample $(\mathbf{x}', \hat{y}) \in \mathcal{S}$. 
We then set a labeled sample as a teacher sample, and set an unlabeled sample as a student sample.
An assistant feature is generated by an assistant generation (AG) module, which transfers an intermediate style of the teacher and the student.
The content of the assistant follows the one of the student.
To blend the style of the pair, we construct new style statistics (\ie, mean and standard deviation) by interpolating between the style statistics of the teacher and the student~\cite{huang2017arbitrary, zhou2021domain}.
Let $f_n(\cdot; \boldsymbol \theta_n)$ be the $n$-th layer of $f(\cdot; \boldsymbol \theta)$ and the intermediate feature $\mathbf{z}_n = f_n(\mathbf{x}; \boldsymbol \theta_n)$, where $\mathbf{z}_n \in R^{C\times H\times W}$.
The style fused feature is calculated as 
\begin{align}
\label{eq:style-fusion}
\text{AG}(\mathbf{z}_n') &= \gamma\frac{\mathbf{z}_n' - \mu(\mathbf{z}_n')}{\sigma(\mathbf{z}_n')} + \beta \text{,} \nonumber\\
\text{where } \beta &= \epsilon\mu(\mathbf{z}_\text{n}) + (1 - \epsilon)\mu(\mathbf{z}_n') \text{,}\nonumber\\ 
\gamma &= \epsilon\sigma(\mathbf{z}_n) + (1 - \epsilon)\sigma(\mathbf{z}_n')\text{,}
\end{align}
where $\epsilon$ is extracted from the Beta distribution $Beta(\rho, \rho)$ with a hyper-parameter $\rho$ and $\mathbf{z}_n'$ is extracted from the student sample $\mathbf{x}'$. We use $\rho = 0.1$ following \cite{zhou2021domain}.
$\mu(\mathbf{z})$, $\sigma(\mathbf{z}) \in R^C$ denote mean and standard deviation of the feature $\mathbf{z}$ across the spatial dimension, respectively:

\vspace{-4mm}
\begin{align}
\label{eq:mean-std}
\mu(\mathbf{z}) &= \frac{1}{HW}\sum_{h=1}^{H}\sum_{w=1}^{W}\mathbf{z}_{hw} \text{,} \\ 
\sigma(\mathbf{z}) &= \sqrt{ \frac{1}{HW}\sum_{h=1}^{H}\sum_{w=1}^{W}(\mathbf{z}_{hw} - \mu(\mathbf{z}))^{2}} \text{.}
\end{align}

AG operation is applied to the intervals of the neural network. See supplementary material for more details.
Note that, unlike \cite{zhou2021domain} that directly trains the model with stylized features (\ie, the gradients are back-propagated through the features), the gradients are blocked when we apply AG module to the intermediate features and generate assistant features.
$f'(\cdot;\boldsymbol{\theta}, \phi)$ is the feature extractor $f$ with AG operation.
Thus the assistant feature is denoted as $f'(\mathbf{x}';\boldsymbol{\theta}, \phi)$, where $\phi = (\boldsymbol{\mu}, \boldsymbol{\sigma}) \in \{\mu(\mathbf{z}_n), \sigma(\mathbf{z}_n)\}_{n=1}^{N}$.
\newline

\noindent
\textbf{Self-distillation and training objectives.}
When implementing a pair loss, we cannot sample all the pairs from two sets, since their cardinalities are large enough. Thus, we uniformly sample the pairs from the set of all possible pairs. The work \cite{LeeJ1990book} shows this sampling technique is an unbiased estimator of true expectation.
The pair loss is calculated as
\begin{eqnarray}
\label{eq:pair_loss}
\mathcal{L}_\text{pair} = \mathbb{E}_{(\mathbf{x},y),(\mathbf{x}',\hat{y})}\!\left[ \llbracket \hat{y} \!=\! y \rrbracket D_\text{KL}(p(\hat{y}|\mathbf{x}',\phi)\|p(\hat{y}|\mathbf{x}'))\right ],
\end{eqnarray}
where $\llbracket \cdot \rrbracket$ denotes Iverson brackets. 
$\phi$ is calculated when the teacher sample $\mathbf{x}$ is forwarded through the feature extractor $f$.
It effectively reduces the inter-domain discrepancy using pairs between $\mathcal{D}_\text{S}$ and $\mathcal{S}$, while suppressing the intra-domain discrepancy using pairs between $\mathcal{D}_\text{TL}$ and $\mathcal{S}$.
This effect is validated in Figure~\ref{fig:hist_all}.

To improve our training, we introduce an additional loss using student samples with pseudo-labels. 
We utilize the latest prediction of student samples to decide the reliability of pseudo-labels and multiply it to the cross-entropy loss of each student sample assuming its pseudo-label as its true label, \ie, we use a weighted cross-entropy loss (WCE) for training student samples:
\begin{eqnarray}
\label{eq:wce_loss}
\mathcal{L}_\text{unl} = \mathbb{E}_{(\mathbf{x}',\hat{y})}[-\omega\log p(\hat{y}|\mathbf{x}')], \quad
\text{where} \, \, \omega = p(\hat{y}|\mathbf{x}').
\end{eqnarray}
By doing so, $\omega$ plays a role in weighting prediction so that less confident samples from $\mathcal{S}$ give less effect on updating the model.

Our total loss in training thus consists of three terms: 
\begin{eqnarray}
\label{overall_loss}
\mathcal{L} = \mathcal{L}_\text{lab} + \mathcal{L}_\text{unl} + \lambda\mathcal{L}_\text{pair},
\end{eqnarray}
where $\lambda$ is a weighting hyper-parameter for the pair loss. 
The model is updated by minimizing $\mathcal{L}$ for $M$ iterations.
We iterate alternating the student-set generation step and the sample pairing, assistant generation and self-distillation step until the model converges on the validation set.

\begin{table*}[t!]
\vspace{2mm}
\begin{center}
\scalebox{0.85}{
\begin{tabular}{l|l|cccccccccccc|c}
\toprule[1pt]
 Net& Method       &R to C& R to P & R to A & P to R & P to C & P to A & A to P & A to C & A to R & C to R & C to A & C to P & MEAN \\ \hline
 \multirow{9}{*}{\rotatebox[origin=c]{90}{AlexNet}} 
&S+T        & 37.5  & 63.1  & 44.8  & 54.3  & 31.7  & 31.5  & 48.8  & 31.1  & 53.3  & 48.5  & 33.9  & 50.8  & 44.1 \\
&DANN       & 42.5  & 64.2  & 45.1  & 56.4  & 36.6  & 32.7  & 43.5  & 34.4  & 51.9  & 51.0  & 33.8  & 49.4  & 45.1 \\
&ADR        & 37.8  & 63.5  & 45.4  & 53.5  & 32.5  & 32.2  & 49.5  & 31.8  & 53.4  & 49.7  & 34.2  & 50.4  & 44.5 \\
&CDAN       & 36.1  & 62.3  & 42.2  & 52.7  & 28.0  & 27.8  & 48.7  & 28.0  & 51.3  & 41.0  & 26.8  & 49.9  & 41.2 \\
&ENT        & 26.8  & 25.8  & 45.8  & 56.3  & 23.5  & 21.9  & 47.4  & 22.1  & 53.4  & 30.8  & 18.1  & 53.6  & 38.8 \\
&MME        & 42.0  & \textbf{69.6}  & 48.3  & 58.7  & \bf{37.8}  & 34.9  & 52.5  & \bf{36.4}  & 57.0  & 54.1  & \bf{39.5}  & 59.1  & 49.2 \\
&APE $\dagger$  & 42.1  & \textbf{69.6}  & \textbf{49.8}  & 57.7  & 35.5  & \textbf{35.9}  & 49.2  & 32.1  & 55.0  & 52.7  & 37.8  & 57.6  & 47.9 \\\cline{2-15}
& \ours w/o AF &   \textbf{45.3}  &  69.5  &   48.0  & 58.5  &   34.8  &   34.5  &  55.9  & 34.6  &   \textbf{57.2}  &   \textbf{56.7}  &   37.0  &   \textbf{60.3}  & 49.4 \\
&\ccol\ours (ours) & \ccol43.0 & \ccol70.1 & \ccol48.4 & \ccol\bf{60.3} & \ccol35.6 & \ccol35.3 & \ccol\bf{56.9} & \ccol35.5 & \ccol56.8 & \ccol55.9 & \ccol37.5 & \ccol59.1 & \ccol\bf{49.5} \\\hline\hline

\multirow{5}{*}{\rotatebox[origin=c]{90}{ResNet34}} 
&S+T & 50.9  & 78.7  & 65.9  & 73.6  & 46.5  & 54.4  & 68.6  & 48.7  & 73.2  & 67.1  & 55.2  & 64.9  & 62.3 \\
&MME & 60.3  & \bf{82.6}  & 71.0  & \bf{79.1}  & \bf{57.9}  & 63.6  & 74.6  & 59.2  & 77.3  & 73.5  & 64.1  & 75.1  & 69.9 \\
&APE $\dagger$    & 60.1  & 82.4  & \bf{73.0}  & 78.5  & 53.3  & 64.6  & 74.7  & 53.4  & 75.7  & 70.6  & 61.6  & 69.1  & 68.1 \\\cline{2-15}
& \ours w/o AF &61.8  & 82.5 &    70.3  & 78.7  & 54.8  &    62.5  &\bf{75.6}  &  58.2  &    \bf{78.3}  &    \bf{74.9}  &     \bf{65.3} &     \bf{77.5} &    70.0 \\
&\ccol\ours (ours) & \ccol\bf{63.2} & \ccol82.3 & \ccol71.0 & \ccol79.0 & \ccol56.8 & \ccol\bf{64.7} & \ccol75.3 & \ccol\bf{59.3} & \ccol77.4 & \ccol73.6 & \ccol64.6 & \ccol76.1 & \ccol\bf{70.3}\\
\bottomrule[1pt]
\end{tabular}
}
\end{center}
\vspace{-4mm}
\caption{Classification accuracy on the Office-Home dataset (\%) for one-shot on four domains (R: Real, C: Clipart, P: Product, A: Art). \label{tb:office_home_all}}
\end{table*}


\section{Experiments}
We compare \ours with current state-of-the-art methods on two standard SSDA benchmarks.
We demonstrate that \ours is generally applicable when there are zero or many target domain labels are available.
Through extensive ablation studies, we verify the effectiveness of each proposed component in detail.
For more experimental setups and results, refer to the supplementary material.
Our code is available on~\url{https://github.com/userb2020/s3d}.

\subsection{Setup}
\noindent
\textbf{Datasets.} We evaluate our method using two benchmark datasets: DomainNet~\cite{peng2019moment} and Office-Home~\cite{venkateswara2017deep}. 
DomainNet contains 6 domains each of which has 345 classes. 
Among them, we use 4 domains (Real, Clipart, Painting, and Sketch) and 126 classes.
We use 145,145 images from all 4 domains for our experiment.
We choose seven source-to-target domain scenarios following the work of \cite{saito2019semi}.
Office-Home consists of 4 domains (Real, Clipart, Product, and Art) of 65 classes, and 15,588 images in total.
We conduct Office-Home experiments on all possible source-to-target domain scenarios. \\

\vspace{-2mm}
\noindent
\textbf{Implementation details.}
We follow most of the implementation details of \cite{saito2019semi} for a fair comparison.
We select AlexNet~\cite{krizhevsky2012imagenet} and ResNet-34~\cite{he2016deep}, both of which are pre-trained on ImageNet~\cite{deng2009imagenet}.
A mini-batch consists of samples from $\mathcal{T}$ and $\mathcal{S}$ at the ratio of 1 to 1.
We choose the same number of source and labeled target data to construct teacher samples in the batch.
Specifically, we use 128 and 96 samples in AlexNet and ResNet-34 respectively as done in MME.
We adopt the SGD optimizer with an initial learning rate of 0.01, a momentum of 0.9, and a weight decay of 0.0005.
In the reliable student-set generation step, we set $\alpha$ to $0.95$.
We set the student-set generation interval $M$ as 100.
We validate the model every 1000 iterations during training, and we early-stop training when the model shows no more improvement in 5 validation steps.
The details of searching the hyper-parameter  $\lambda$ are described in the supplementary material.
We use PyTorch \cite{paszke2017automatic} for our experiments.

\noindent
\textbf{Baselines.} We compare our method to competitive SSDA baselines: MME \cite{saito2019semi}, and APE \cite{kim2020attract}. Additionally, we bring S+T that simply minimizes the cross-entropy loss on the labeled dataset. 
\ours w/o AF is our method without the assistant, \ie directly distills the teacher prediction to the student.
DANN~\cite{ganin2016domain}, ADR~\cite{saito2017adversarial}, and CDAN ~\cite{long2018conditional}, which are the well-known methods in UDA, are also described as a comparison. Further, we include the accuracy of ENT~\cite{grandvalet2005semi}.

\subsection{Main results}
We conduct experiments on both one-shot and three-shot settings with AlexNet and ResNet-34.
For a fair comparison, we select the best model on the validation set for all experiments. \\

\noindent
\textbf{Comparison on DomainNet.} 
Table~\ref{tb:domain_all} compares the classification accuracy of our method and other baselines on the DomainNet dataset.
\ours achieves higher accuracy than the previous methods in most domain adaptation scenarios.
Notably, \ours is effective where the domain gap between the source and target domain is substantial.
In comparison to S+T, for example, \ours increases accuracy by 17.7\%p on Real to Clipart with ResNet in the one-shot setting.
Real and Clipart domains appear considerably distinctive to each other because samples in Real domain are photos, whereas, the samples in Clipart domain are artificial illustrations.
The mixed style assistant seems to be effective when comparing the accuracy between \ours and \ours w/o AF.\\

\begin{figure}[ht!]
  \begin{subfigure}{0.23\textwidth}
    \includegraphics[width=\linewidth]{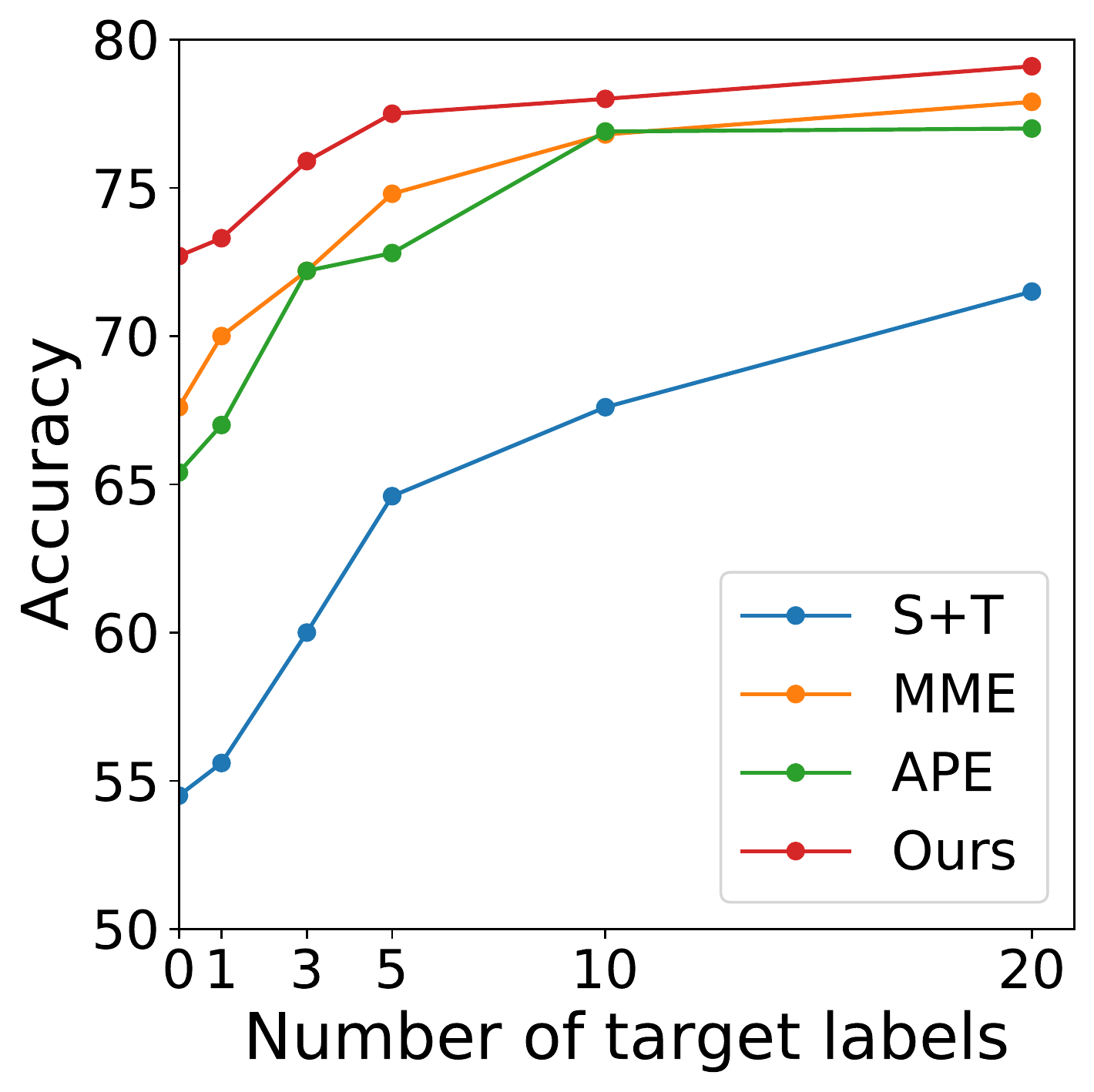}
    \vspace{-4mm}
    \caption{Real to Clipart. \label{fig:many_r_to_c2}}
  \end{subfigure}
  \hspace*{\fill} 
  \begin{subfigure}{0.23\textwidth}
    \includegraphics[width=\linewidth]{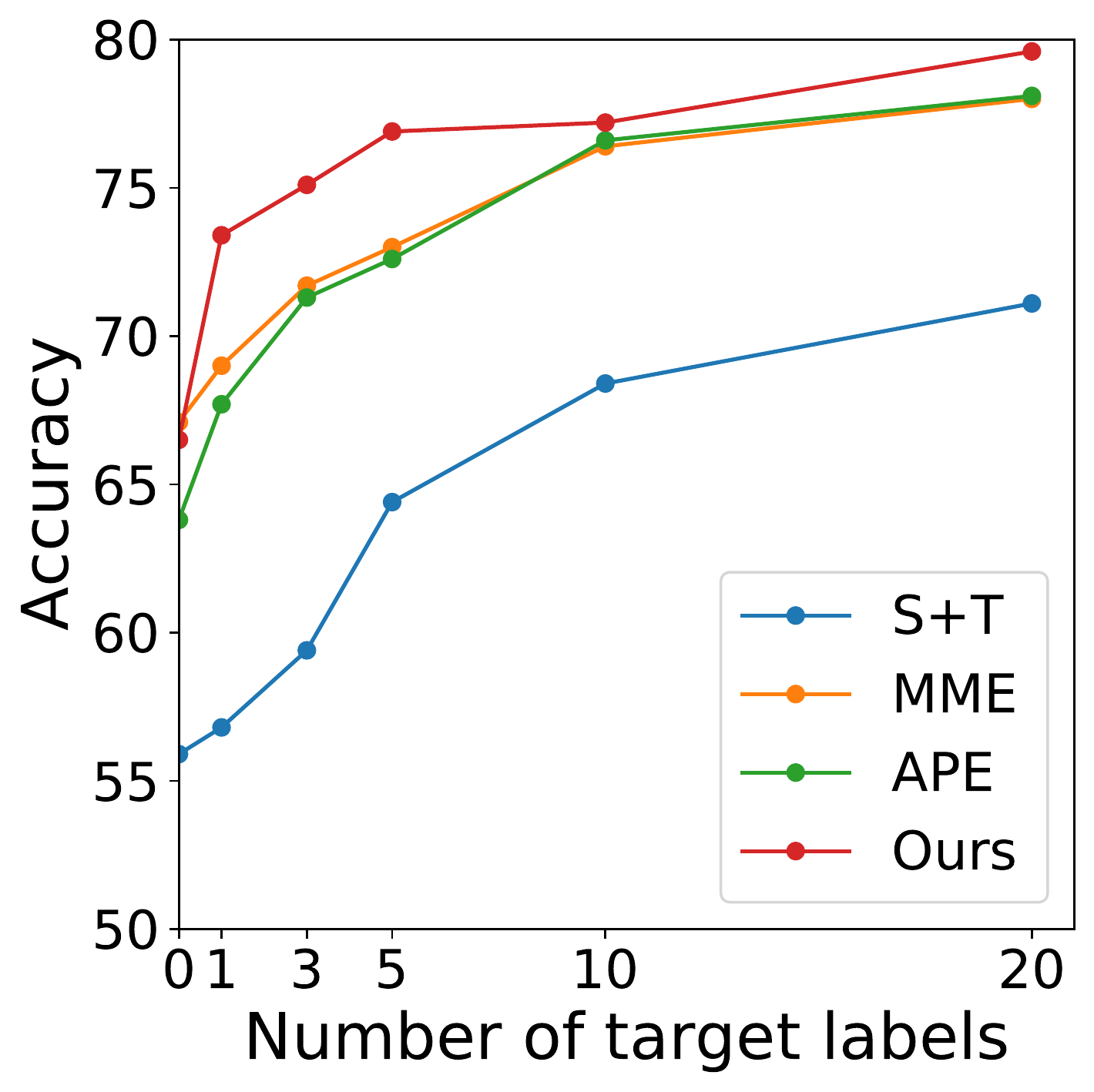}
    \vspace{-4mm}
    \caption{Painting to Clipart. \label{fig:many_p_to_c2}}
  \end{subfigure}
\caption{Classification accuracy (\%) of two scenarios on DomainNet with increasing number of target labels. \label{fig:many_shot}}
\vspace{-2mm}
\end{figure}

\begin{table}[t!]
\begin{center}
\scalebox{0.75}{
\begin{tabular}{lcccccccc}
\toprule[1pt]
Method       & R-C & R-P & P-C & C-S & S-P & R-S & P-R & MEAN \\ \midrule
&\multicolumn{7}{c}{AlexNet} &\\ \midrule
Source       & 41.1   & 42.6   & 37.4   & 30.6   & 30.0   & 26.3   & 52.3   & 37.2 \\
DANN         & 44.7   & 36.1   & 35.8   & 33.8   & 35.9   & 27.6   & 49.3   & 37.6 \\
ADR          & 40.2   & 40.1   & 36.7   & 29.9   & 30.6   & 25.9   & 51.5   & 36.4 \\
CDAN         & 44.2   & 39.1   & 37.8   & 26.2   & 24.8   & 24.3   & 54.6   & 35.9 \\
ENT          & 33.8   & 43.0   & 23.0   & 22.9   & 13.9   & 12.0   & 51.2   & 28.5 \\
MME          & 47.6   & 44.7   & 39.9   & 34.0   & 33.0   & 29.0   & 53.5   & 40.2 \\
APE $\dagger$  & 45.9   & 47.0   & 42.0   & 36.5   & 37.0   & 30.3   & 54.1   & 41.8 \\\hline
\ours w/o AF & 49.3   &  49.2   &  42.7   &   38.1   &  41.7   & \textbf{38.0}   &   54.1   & 44.7 \\ 
\ccol\ours (ours) & \ccol\bf{53.4} & \ccol\bf{51.9} & \ccol\bf{46.3} & \ccol\bf{38.7} & \ccol\bf{44.0} & \ccol36.4 & \ccol\bf{57.6} & \ccol\bf{46.9}\\  \midrule
&\multicolumn{7}{c}{ResNet34} &\\ \midrule
Source       & 54.5  & 60.2  & 55.9 & 49.7 & 50.1 & 44.1 & 72.1 & 55.2 \\
MME          & 67.6  & 66.9  & \bf{67.1}   & 56.4  & 62.9 & 58.2  & 74.5  & 64.8 \\
APE $\dagger$ & 65.4 & 68.6 & 63.8  & 56.4 & \bf{65.1} & 60.4  & \bf{75.3}  & 65.0 \\\hline
\ours w/o AF & 70.1 &  69.5 &  66.8  &  56.3   &    61.8 &  59.0  &    73.4  &  65.3 \\
 \ccol\ours (ours) &\ccol \bf{72.7} & \ccol\bf{70.2} & \ccol66.5 & \ccol\bf{57.2} & \ccol63.8 & \ccol\bf{62.6} & \ccol71.2 & \ccol\bf{66.3} \\
\bottomrule[1pt]
\end{tabular}
}
\end{center}
\vspace{-5mm}
\caption{Classification accuracy on the DomainNet dataset (\%) in the unsupervised domain adaptation setting. \label{tb:uda}}
\vspace{-3mm}
\end{table}

\noindent
\textbf{Comparison on Office-Home.}
Table~\ref{tb:office_home_all} compares the results of our method and others on the Office-Home dataset.
Our method shows comparable results with other baselines. 
As \ours in DomainNet, the method is powerful in adapting to the quite different domain, for example, on Real to Clipart and Art to Product scenarios.
However, the accuracy enhancement of \ours on Office-Home is not as high as that on DomainNet.
We suppose that the small dataset size of the Office-Home derives less performance improvement compared to that on the DomainNet dataset.
Note that the dataset size of the DomainNet is ten times larger than that of the Office-Home.
Considering that student samples acquire diverse guidance from source and labeled target dataset using our method, a larger dataset is more advantageous for \ours to give rich guidance to unlabeled targets.

\begin{figure*}[t!]
  \begin{subfigure}{0.23\textwidth}
    \includegraphics[width=\linewidth]{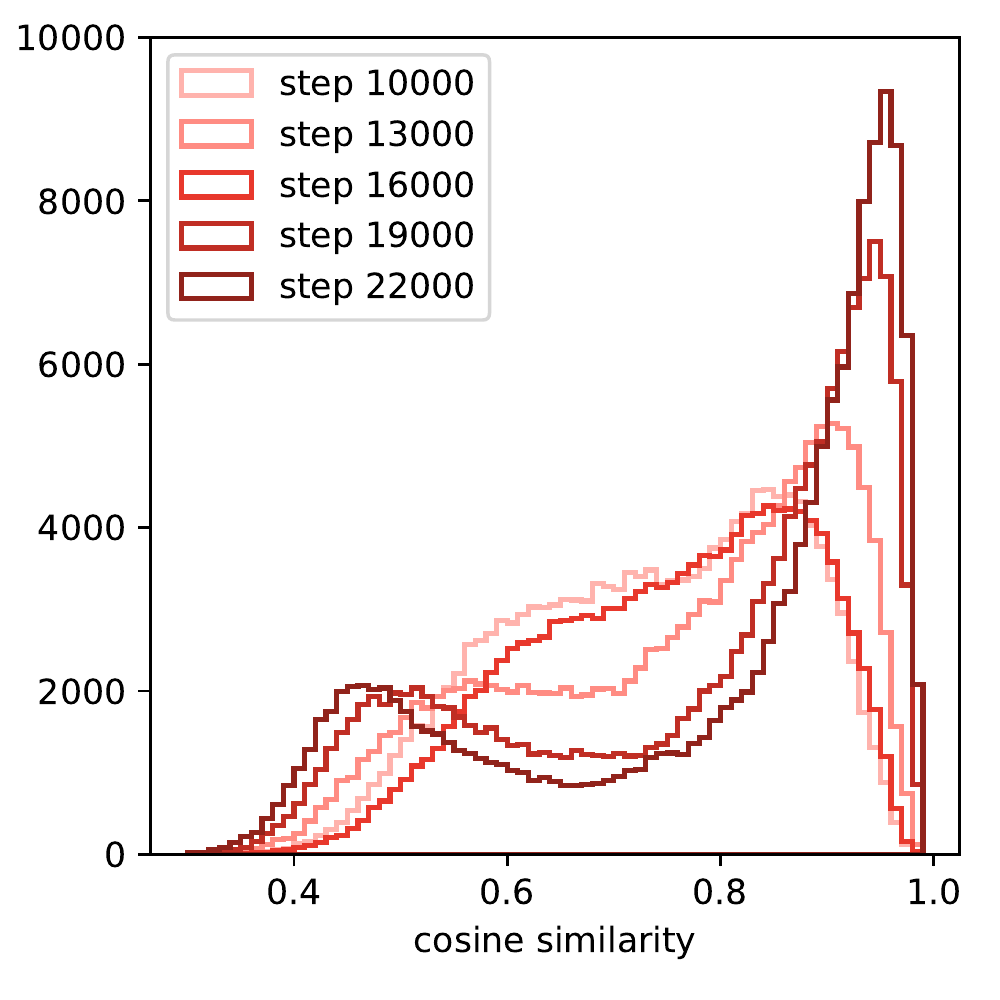}
    \vspace{-4mm}
    \caption{\label{fig:inter}}
  \end{subfigure}
  \hspace*{\fill}   
  \begin{subfigure}{0.23\textwidth}
    \includegraphics[width=\linewidth]{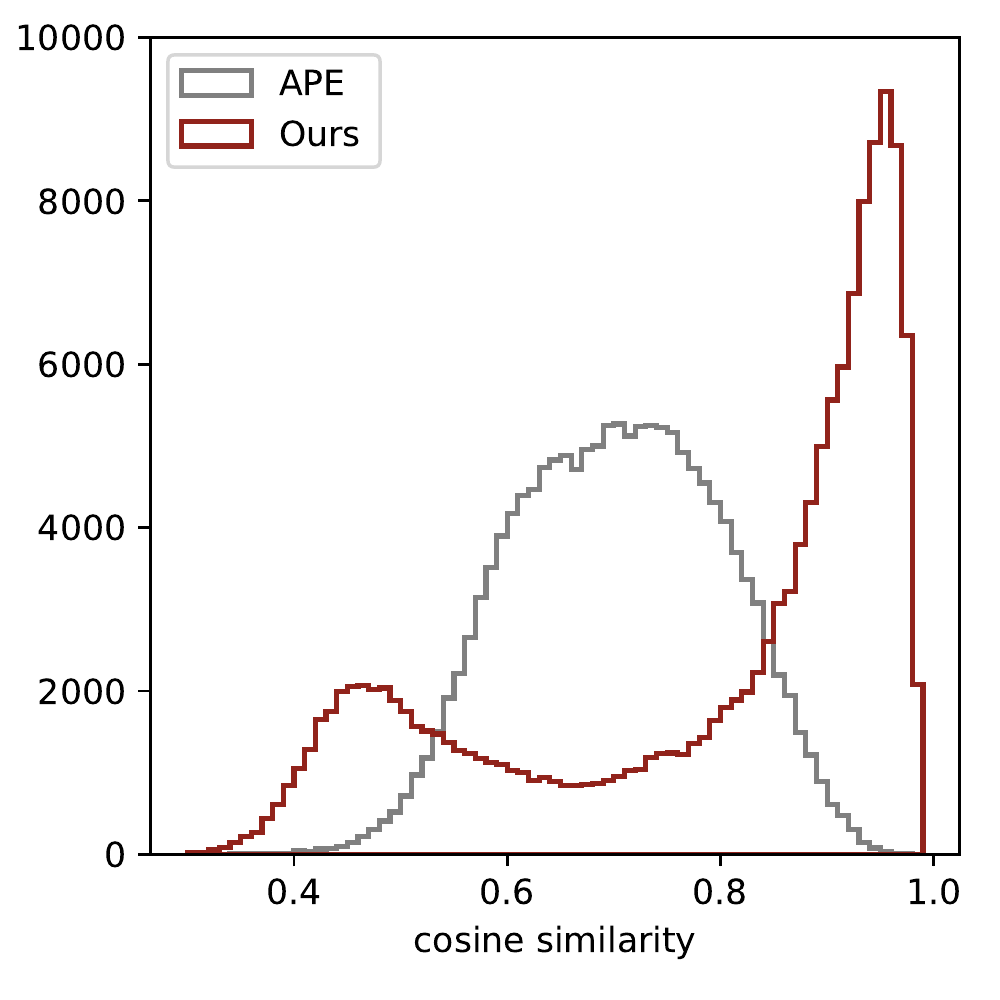}
    \vspace{-4mm}
    \caption{\label{fig:two_inter2}}
  \end{subfigure}
  \hspace*{\fill}   
  \hspace{4mm}
  \begin{subfigure}{0.23\textwidth}
    \includegraphics[width=\linewidth]{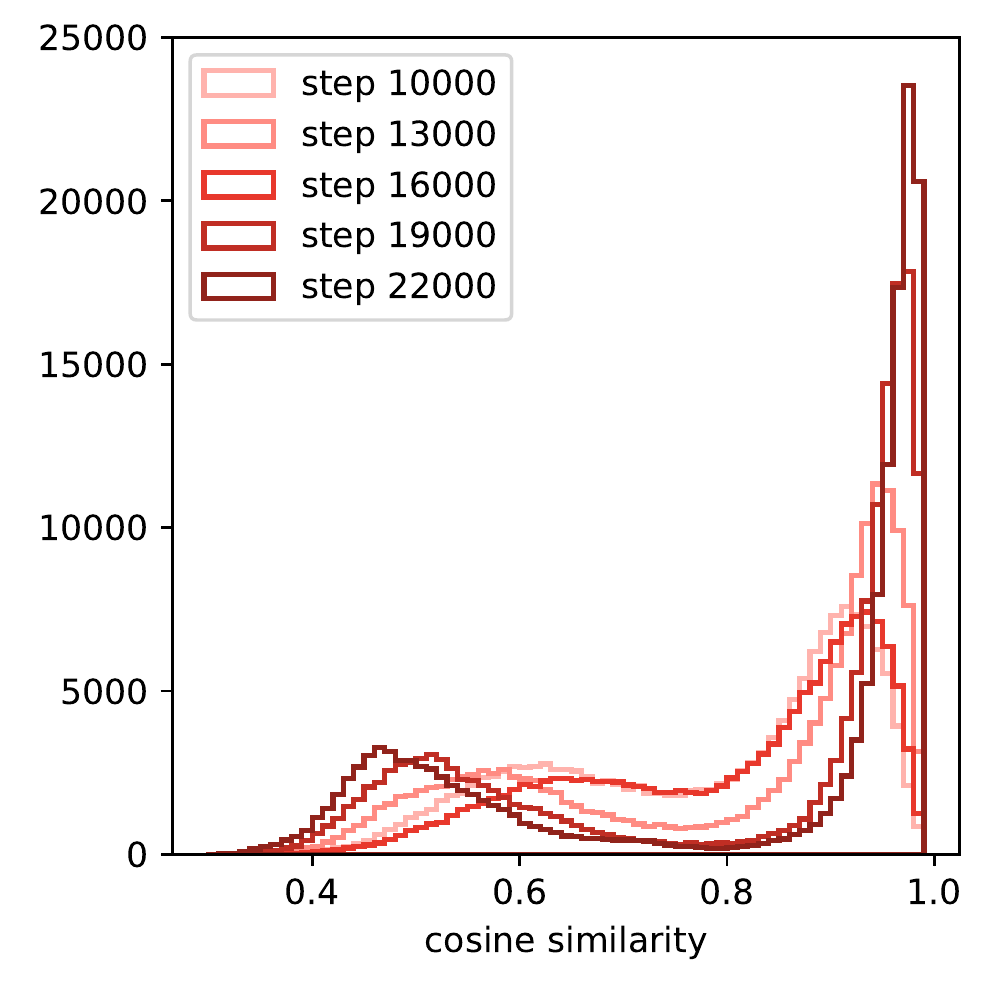}
    \vspace{-4mm}
    \caption{\label{fig:intra}}
  \end{subfigure}
  \hspace*{\fill}   
    \begin{subfigure}{0.23\textwidth}
    \includegraphics[width=\linewidth]{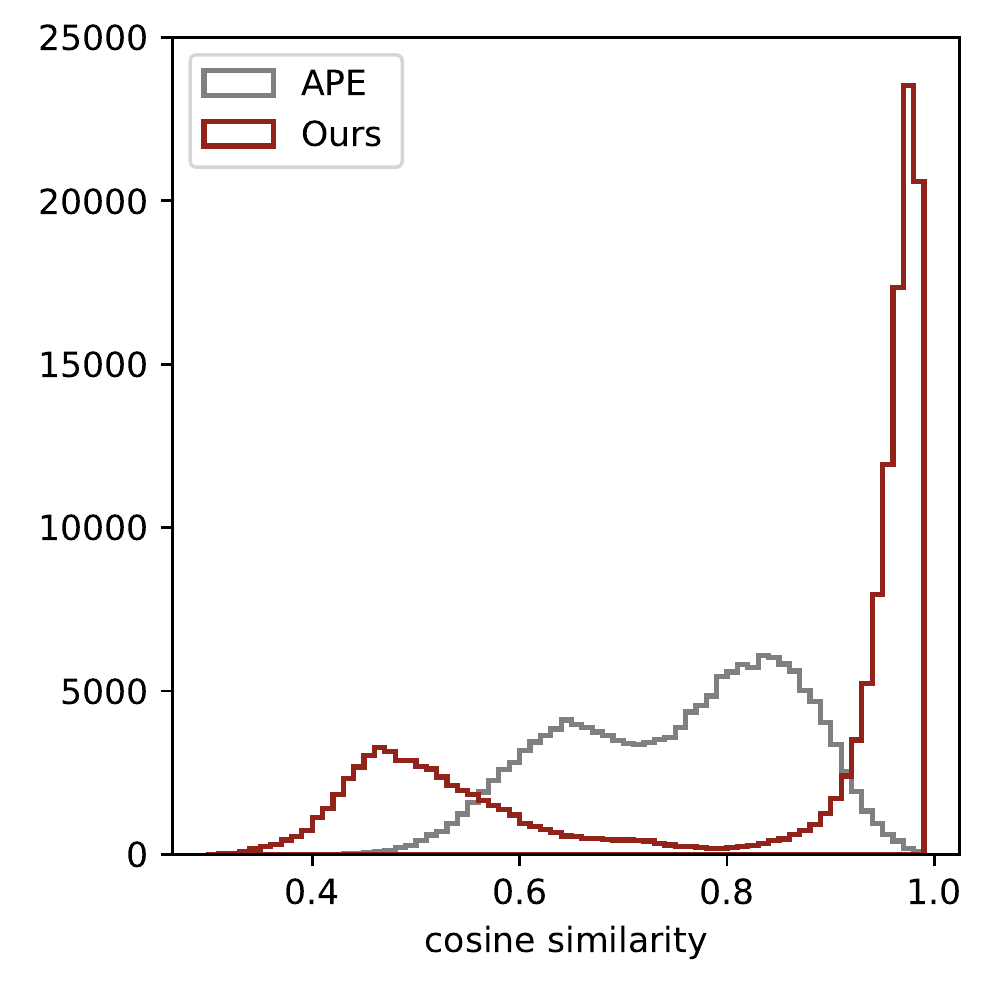}
    \vspace{-4mm}
    \caption{\label{fig:two_intra2}}
  \end{subfigure}
\vspace{-4mm}
\caption{
(a) Histograms of cosine similarity between a source and a target embedding (inter-domain similarity) from the pre-training stage until convergence.
(b) Inter-domain similarity histograms of APE and \ours.
(c) Histograms of cosine similarity between target embeddings (intra-domain similarity) from the pre-training stage until convergence.
(d) Intra-domain similarity histograms of APE and \ours. \label{fig:hist_all}}
\end{figure*}

\begin{figure}[t!]
  \begin{subfigure}{0.23\textwidth}
    \includegraphics[width=\linewidth]{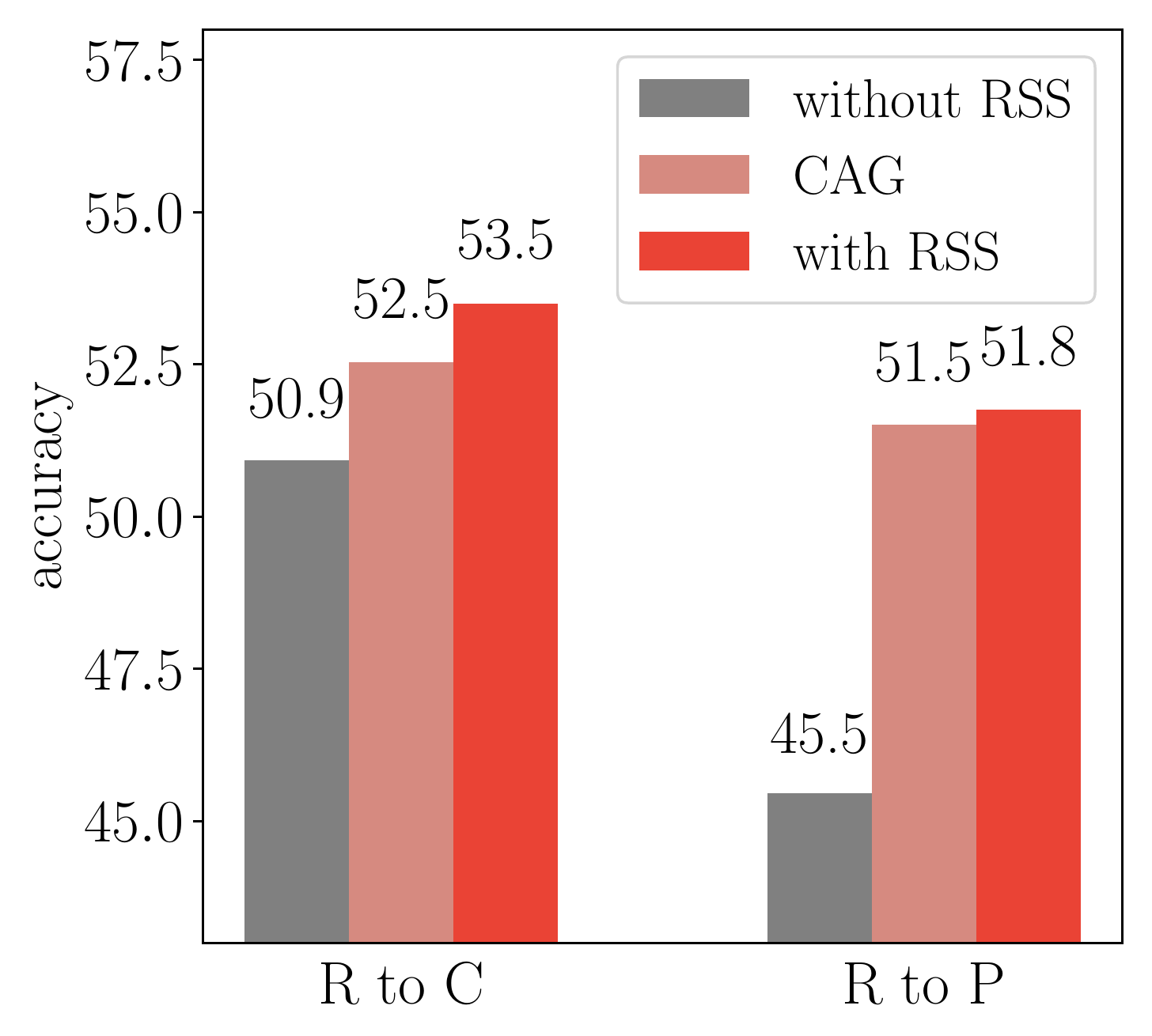}
    \vspace{-6mm}
    \caption{Ablation of RSS. \label{fig:rss_ablation}}
  \end{subfigure}
  \hspace*{\fill}  
  \begin{subfigure}{0.23\textwidth}
    \includegraphics[width=\linewidth]{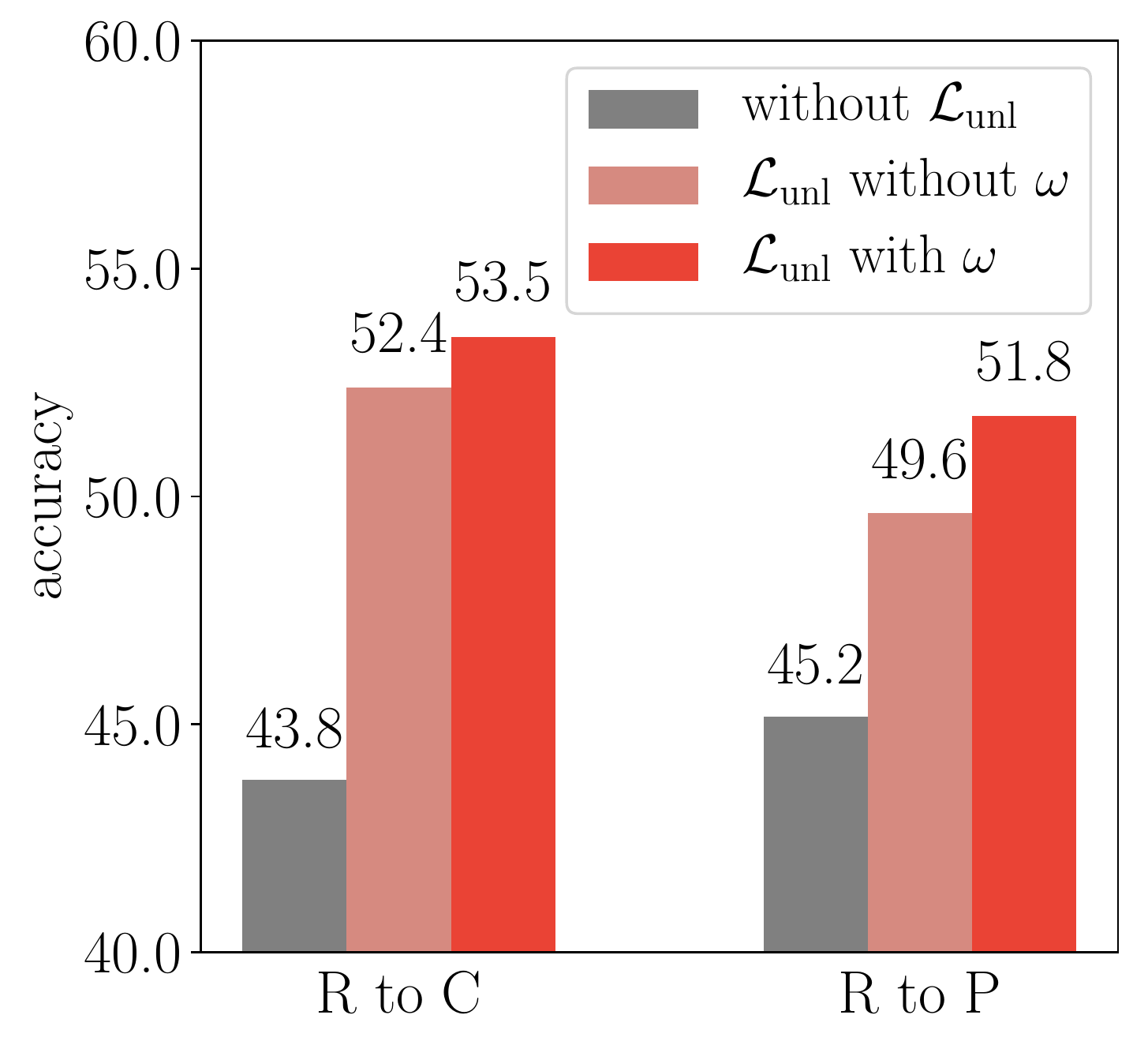}
    \vspace{-6mm}
    \caption{Ablation of $\mathcal{L}_{\text{unl}}$. \label{fig:wce_ablation}}
  \end{subfigure}
\caption{
Ablation study of proposed components. \label{fig:ablation_components}}
\vspace{-4mm}
\end{figure}

\subsection{Varying the number of target labels}

\noindent
\textbf{Many-shot semi-supervised domain adaptation.} We examine our method with increasing target labels per class.
Figure~\ref{fig:many_shot} reports the many-shot experiment results.
\ours consistently outperforms current state-of-the-art along with the increasing number of target domain ground truths.
Note that the S+T baseline corresponds to the pre-training stage in our context.
This experiment emphasizes the strength of the sample-to-sample training even when abundant target labels are given in the pre-training stage.
In the 20-shot case, for example, the pre-training is supervised by 2,520 target domain examples, and yet the following sample-to-sample training stage gains additional accuracy improvement of 8.5\%p from the pre-trained model.\\

\noindent
\textbf{Unsupervised domain adaptation.}
In Table~\ref{tb:uda}, \ours is also shown to be effective even though no target ground truth is available.
For UDA experiments, we compose training batch samples from the source dataset and the unlabeled target dataset, and set validation batches as the same as the one used in the semi-supervised setup.
\ours outperforms counterparts in most scenarios.
It is impressive that \ours excels methods that have been proposed for UDA~\cite{ganin2016domain, saito2017adversarial, long2018conditional}, each of which involves a domain-adversarial learning objective.

\subsection{Ablation study}
\noindent
\textbf{Ablation study on proposed components.} We conduct extensive ablation studies on our major contributions: the pair loss $\mathcal{L}_\text{pair}$, the weighted-cross entropy loss $\mathcal{L}_\text{unl}$ and the reliable student-set generation (RSS) in Table~\ref{table:student}.
The top row of Table~\ref{table:student} represents the pre-training stage which is equivalent to the S+T baseline.
It attempts to train a network only with labeled samples without any proposed components.
The check mark $\checkmark$ on the RSS column indicates that we filter out unreliable pseudo-labeled samples using \Eqref{eq:reliability_evaluation}, otherwise we utilize all pseudo-labeled samples into training. 

As can be seen in Table~\ref{table:student}, the pair loss contributes to performance improvement.
Using the pair loss and the weighted cross-entropy together significantly increases the accuracy by a large margin from the top-row baseline on all scenarios.
Additionally, RSS further improves performance by preventing harmful alignment from mis-matched pairs.
The bottom row completes our method, and it validates that all components are complementary. \\

\noindent
\textbf{Ablation study on evaluating reliability of pseudo-labels.} We examine that discarding unreliable pseudo-labels is crucial in training with pseudo-labels.
We compare our RSS with other pseudo-label reliability evaluation scheme.
In Figure~\ref{fig:rss_ablation}, \textsl{without RSS} indicates that we set all pseudo-labeled samples to a student set regardless of their reliability, and \textsl{CAG}~\cite{zhang2019category} indicates that we measure the reliability using the first condition of \Eqref{eq:reliability_evaluation}. 
While CAG~\cite{zhang2019category} searches for an optimal margin $\delta$, our model sets the $\delta$ to the average margin value of all student set thus eliminating a hyper-parameter.
Our RSS method effectively constructs reliable pairs thus improving performance over two baselines.
Compared to the performance gain from $\mathcal{L}_{\text{unl}}$, which will be reported in the following paragraph with Figure~\ref{fig:wce_ablation}, the gain of reliable paring is clearer.
This comparison witnesses again that the pair-based learning brings the major performance improvement in our pipeline. \\

\noindent
\textbf{Ablation study on weighted cross-entropy.} In Figure~\ref{fig:wce_ablation}, we verify the effect of the confidence score $\omega$ in the weighted cross-entropy of \Eqref{eq:wce_loss}. 
In Figure~\ref{fig:wce_ablation}, \textsl{without} $\mathcal{L}_{\text{unl}}$ indicates that we exclude $\mathcal{L}_{\text{unl}}$ in the overall loss term, and  $\mathcal{L}_{\text{unl}}$ \textsl{without} $\omega$ indicates that we eliminate $\omega$ in $\mathcal{L}_{\text{unl}}$.
We show that multiplying $\omega$ to the cross-entropy term achieves additional performance growth. \\

\begin{table}[t!]
\vspace{4mm}
\centering
\scalebox{0.70}{
\begin{tabular}{ccc|cccccccc}
\toprule[1pt]
$\mathcal{L}_\text{unl}$ & $\mathcal{L}_\text{pair}$ & RSS & R-C & R-P & P-C & C-S & S-P & R-S & P-R & MEAN\\ \hline
\xmark & \xmark & \xmark &56.8  & 60.5 & 55.4 & 51.7 & 55.5 & 47.5 & 72.0 & 57.1\\
\cmark & \xmark & \xmark & 68.7 & 65.6 & 68.8 & 59.2 & 64.1 & 61.6 & 78.4 & 66.6 \\
 \xmark &\cmark& \xmark & 67.4 & 65.0 & 67.1 & 61.2 & 64.9 & 62.7 & 77.5 & 66.5 \\
\cmark &\cmark& \xmark & 69.4 & 65.7 & 69.7 & 61.3 & 65.5 & 61.7 & 78.6 & 67.4 \\
 \ccol \cmark & \ccol \cmark& \ccol \cmark & \ccol 73.3  & \ccol 68.9 & \ccol 73.4 & \ccol 60.8 & \ccol 68.2 & \ccol 65.1 & \ccol 79.5 & \ccol 69.9  \\
\bottomrule[1pt]
\end{tabular}}
\caption{Comprehensive ablation study of \ours on the DomainNet dataset for one-shot setting. \label{table:student} }
\vspace{-8mm}
\end{table}

\vspace{-4mm}
\subsection{Analysis}
\noindent
\textbf{Inter-domain and intra-domain discrepancies.}
\label{sec:discrepancies}
Figure~\ref{fig:hist_all} visualizes that \ours progressively clusters instances of the same classes by overcoming inter- and intra-domain discrepancies.
Figure~\ref{fig:inter} plots a histogram of cosine similarity between a source and a target embedding from the same class for all classes.
Figure~\ref{fig:intra} plots the one between labeled and unlabeled target embeddings.
Figures~\ref{fig:two_inter2} and \ref{fig:two_intra2} visualize cosine similarity histograms from the final model of APE~\cite{kim2020attract} and \ours.
The similarity population gradually moves toward 1.0 over iterations, which proves that method guides to map two semantically similar samples to nearby points in the embedding space.
While a majority of the same-class embeddings moves close to each other, we observe that a small portion of embeddings pushes apart.
This is considered one limitation of leveraging pseudo-labels; wrong pairs misguide the learning process.\\

\begin{figure}[t!]
 \begin{subfigure}{0.23\textwidth}
    \includegraphics[width=\linewidth]{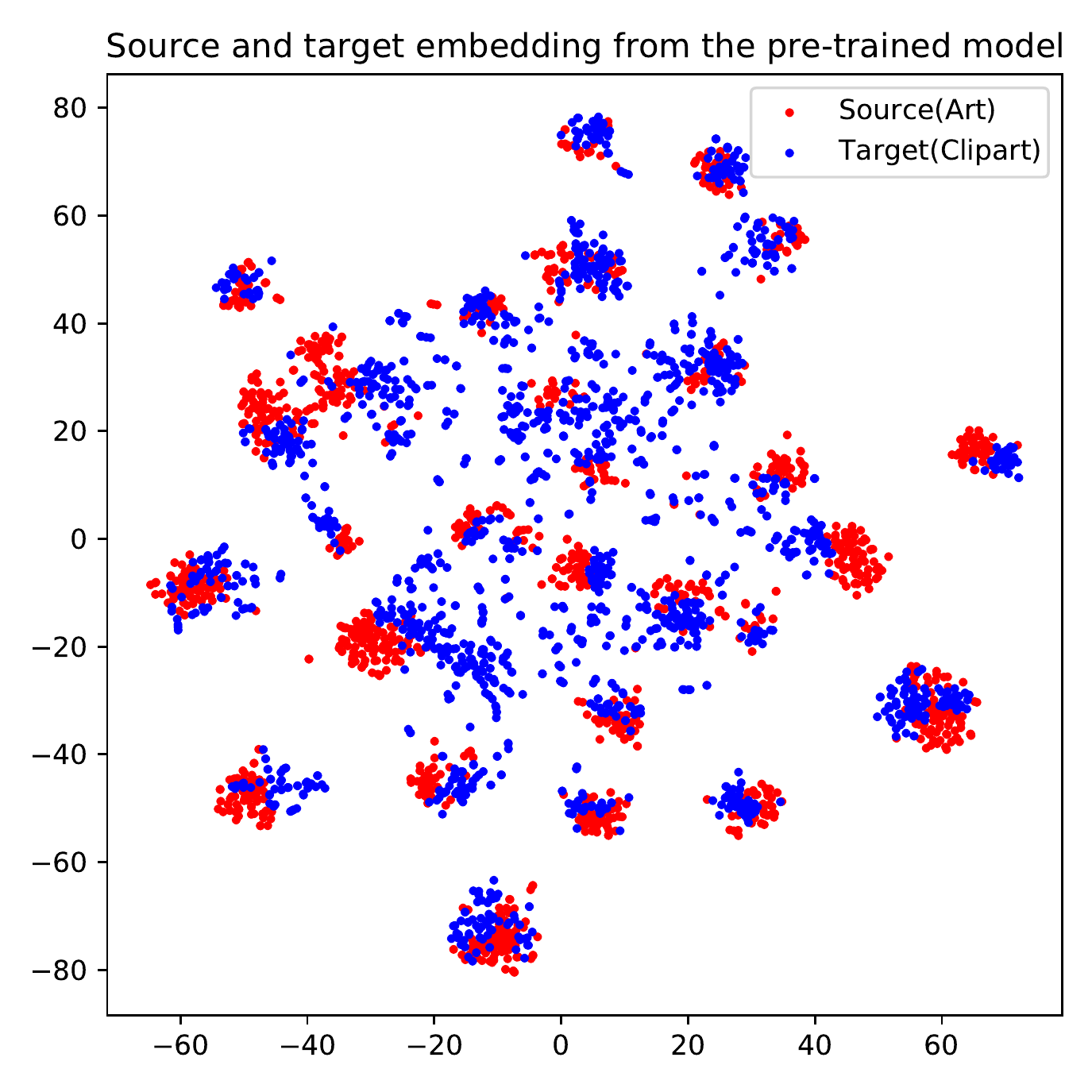}
    \vspace{-4mm}
    \caption{\label{fig:tsne_ac_pre_all}}
 \end{subfigure}
 \hspace*{\fill}   
 \begin{subfigure}{0.23\textwidth}
    \includegraphics[width=\linewidth]{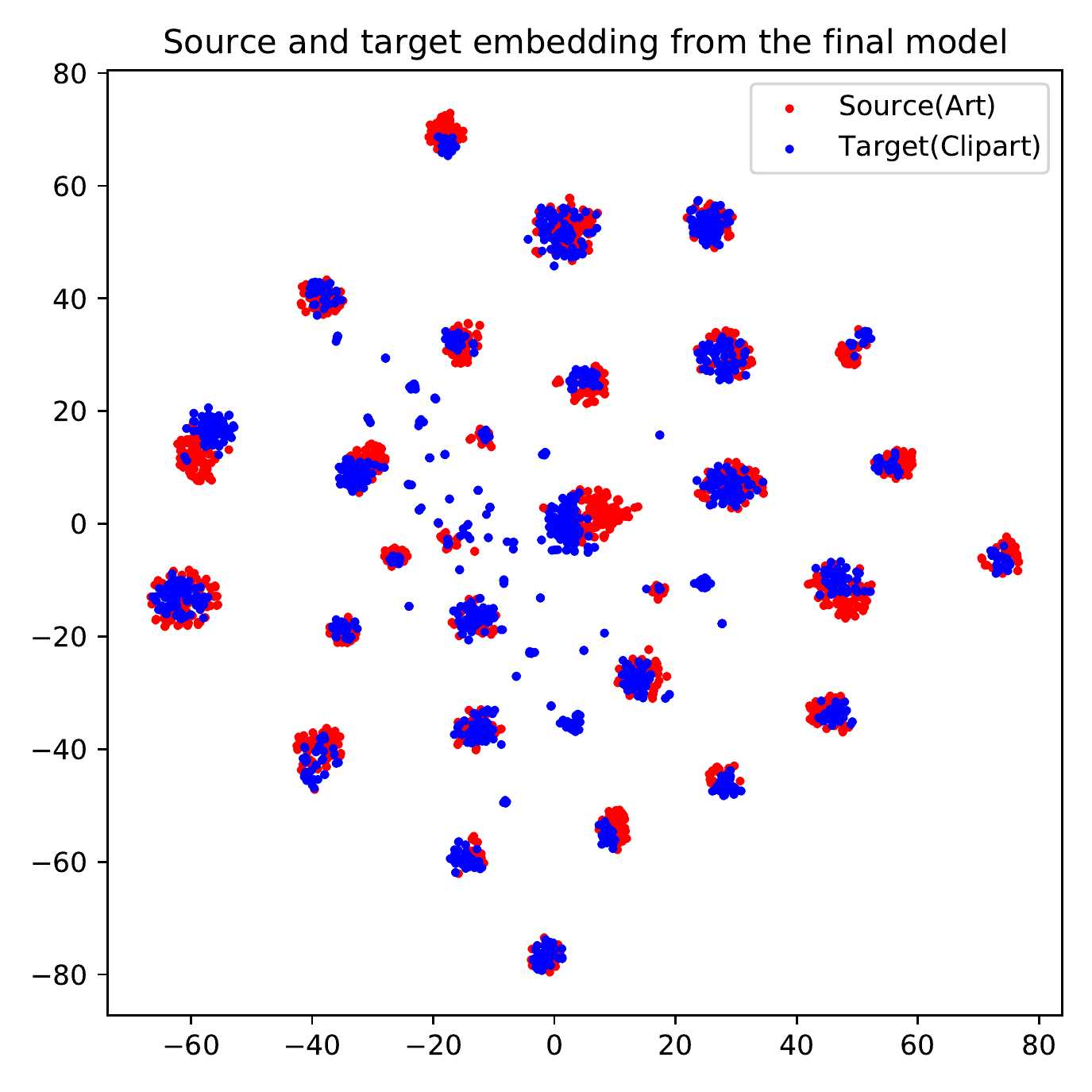}
    \vspace{-4mm}
    \caption{\label{fig:tsne_ac_final_all}}
 \end{subfigure}
\vspace{-2mm}
\caption{
$t$-SNE visualization on the A-C scenario in the Office-Home dataset.
(a) Embedding space from the pre-trained model. 
(b) Embedding space from the final model.
\label{fig:tsne_all}}
\vspace{-5mm}
\end{figure}

\vspace{-3mm}
\noindent
\textbf{Embedding space visualization.}
Figure~\ref{fig:tsne_all} visualizes how \ours clusters instances from two domains over iterations using $t$-SNE~\cite{maaten2008visualizing}.
We observe that \ours clearly enhances the embedding quality from the pre-training stage.
We include more qualitative results in the supplementary material.

\vspace{-1mm}
\section{Conclusion}
We propose a novel sample-to-sample self-distillation (\ours) by exploiting rich and diverse relations for semi-supervised domain adaptation.
By exploiting an assistant feature, the style of which is mixed, \ours efficiently reduce the domain shift.
The experiments demonstrate that \ours effectively adapts to a target domain using a single architecture given an extremely few number of labeled target domain samples. \\
\vspace{-9mm}
\noindent
\paragraph{Acknowledgements.}
This work was supported by Samsung Electronics Co., Ltd. (IO201208-07822-01) and the IITP grants (No.2019-0-01906, AI Graduate School Program - POSTECH) (No.2021-0-00537, Visual common sense through self-supervised learning for restoration of invisible parts in images) funded by Ministry of Science and ICT, Korea.
\newpage

{\small
\bibliographystyle{ieee_fullname}
\bibliography{egbib}

\begin{thebibliography}{10}\itemsep=-1pt

\bibitem{ao2017fast}
Shuang Ao, Xiang Li, and Charles~X Ling.
\newblock Fast generalized distillation for semi-supervised domain adaptation.
\newblock In {\em Proceedings of the Thirty-First AAAI Conference on Artificial
  Intelligence}, pages 1719--1725, 2017.

\bibitem{breiman1996born}
Leo Breiman and Nong Shang.
\newblock Born again trees.
\newblock {\em University of California, Berkeley, Berkeley, CA, Technical
  Report}, 1:2, 1996.

\bibitem{bucilua2006model}
Cristian Bucilu\v{a}, Rich Caruana, and Alexandru Niculescu-Mizil.
\newblock Model compression.
\newblock In {\em Proceedings of the 12th ACM SIGKDD International Conference
  on Knowledge Discovery and Data Mining}, KDD '06, page 535–541, New York,
  NY, USA, 2006. Association for Computing Machinery.

\bibitem{chang2019domain}
Woong-Gi Chang, Tackgeun You, Seonguk Seo, Suha Kwak, and Bohyung Han.
\newblock Domain-specific batch normalization for unsupervised domain
  adaptation.
\newblock In {\em Proceedings of the IEEE Conference on Computer Vision and
  Pattern Recognition}, pages 7354--7362, 2019.

\bibitem{chen2019closer}
Wei-Yu Chen, Yen-Cheng Liu, Zsolt Kira, Yu-Chiang Wang, and Jia-Bin Huang.
\newblock A closer look at few-shot classification.
\newblock In {\em International Conference on Learning Representations}, 2019.

\bibitem{cheng2020explaining}
Xu Cheng, Zhefan Rao, Yilan Chen, and Quanshi Zhang.
\newblock Explaining knowledge distillation by quantifying the knowledge.
\newblock In {\em Proceedings of the IEEE/CVF Conference on Computer Vision and
  Pattern Recognition}, pages 12925--12935, 2020.

\bibitem{cordts2016cityscapes}
Marius Cordts, Mohamed Omran, Sebastian Ramos, Timo Rehfeld, Markus Enzweiler,
  Rodrigo Benenson, Uwe Franke, Stefan Roth, and Bernt Schiele.
\newblock The cityscapes dataset for semantic urban scene understanding.
\newblock In {\em Proceedings of the IEEE conference on computer vision and
  pattern recognition}, pages 3213--3223, 2016.

\bibitem{deng2009imagenet}
Jia Deng, Wei Dong, Richard Socher, Li-Jia Li, Kai Li, and Li Fei-Fei.
\newblock Imagenet: A large-scale hierarchical image database.
\newblock In {\em 2009 IEEE conference on computer vision and pattern
  recognition}, pages 248--255. Ieee, 2009.

\bibitem{deng2019cluster}
Zhijie Deng, Yucen Luo, and Jun Zhu.
\newblock Cluster alignment with a teacher for unsupervised domain adaptation.
\newblock In {\em Proceedings of the IEEE International Conference on Computer
  Vision}, pages 9944--9953, 2019.

\bibitem{donahue2013semi}
Jeff Donahue, Judy Hoffman, Erik Rodner, Kate Saenko, and Trevor Darrell.
\newblock Semi-supervised domain adaptation with instance constraints.
\newblock In {\em Proceedings of the IEEE conference on computer vision and
  pattern recognition}, pages 668--675, 2013.

\bibitem{furlanello2018born}
Tommaso Furlanello, Zachary~C Lipton, Michael Tschannen, Laurent Itti, and
  Anima Anandkumar.
\newblock Born again neural networks.
\newblock {\em arXiv preprint arXiv:1805.04770}, 2018.

\bibitem{ganin2016domain}
Yaroslav Ganin, Evgeniya Ustinova, Hana Ajakan, Pascal Germain, Hugo
  Larochelle, Fran{\c{c}}ois Laviolette, Mario Marchand, and Victor Lempitsky.
\newblock Domain-adversarial training of neural networks.
\newblock {\em The Journal of Machine Learning Research}, 17(1):2096--2030,
  2016.

\bibitem{gong2019dlow}
Rui Gong, Wen Li, Yuhua Chen, and Luc~Van Gool.
\newblock Dlow: Domain flow for adaptation and generalization.
\newblock In {\em Proceedings of the IEEE/CVF Conference on Computer Vision and
  Pattern Recognition}, pages 2477--2486, 2019.

\bibitem{grandvalet2005semi}
Yves Grandvalet and Yoshua Bengio.
\newblock Semi-supervised learning by entropy minimization.
\newblock In {\em Advances in neural information processing systems}, pages
  529--536, 2005.

\bibitem{he2016deep}
Kaiming He, Xiangyu Zhang, Shaoqing Ren, and Jian Sun.
\newblock Deep residual learning for image recognition.
\newblock In {\em Proceedings of the IEEE conference on computer vision and
  pattern recognition}, pages 770--778, 2016.

\bibitem{hinton2015distilling}
Geoffrey Hinton, Oriol Vinyals, and Jeff Dean.
\newblock Distilling the knowledge in a neural network.
\newblock {\em arXiv preprint arXiv:1503.02531}, 2015.

\bibitem{hoffman2018cycada}
Judy Hoffman, Eric Tzeng, Taesung Park, Jun-Yan Zhu, Phillip Isola, Kate
  Saenko, Alexei Efros, and Trevor Darrell.
\newblock Cycada: Cycle-consistent adversarial domain adaptation.
\newblock In {\em International conference on machine learning}, pages
  1989--1998, 2018.

\bibitem{huang2017arbitrary}
Xun Huang and Serge Belongie.
\newblock Arbitrary style transfer in real-time with adaptive instance
  normalization.
\newblock In {\em Proceedings of the IEEE International Conference on Computer
  Vision}, pages 1501--1510, 2017.

\bibitem{jiangbidirectional}
Pin Jiang, Aming Wu, Yahong Han, Yunfeng Shao, Meiyu Qi, and Bingshuai Li.
\newblock Bidirectional adversarial training for semi-supervised domain
  adaptation.
\newblock {\em Twenty-Ninth International Joint Conference on Artificial
  Intelligence}, 2020.

\bibitem{kim2020attract}
Taekyung Kim and Changick Kim.
\newblock Attract, perturb, and explore: Learning a feature alignment network
  for semi-supervised domain adaptation.
\newblock In {\em European conference on computer vision}, 2020.

\bibitem{krizhevsky2012imagenet}
Alex Krizhevsky, Ilya Sutskever, and Geoffrey~E Hinton.
\newblock Imagenet classification with deep convolutional neural networks.
\newblock In {\em Advances in neural information processing systems}, pages
  1097--1105, 2012.

\bibitem{lee2013pseudo}
Dong-Hyun Lee.
\newblock Pseudo-label: The simple and efficient semi-supervised learning
  method for deep neural networks.
\newblock In {\em Workshop on challenges in representation learning, ICML},
  volume~3, 2013.

\bibitem{LeeJ1990book}
Justin Lee.
\newblock {\em U-statistics: Theory and Practice}.
\newblock Marcel Dekker, Inc., New York, 1990.

\bibitem{lee2019drop}
Seungmin Lee, Dongwan Kim, Namil Kim, and Seong-Gyun Jeong.
\newblock Drop to adapt: Learning discriminative features for unsupervised
  domain adaptation.
\newblock In {\em Proceedings of the IEEE International Conference on Computer
  Vision}, pages 91--100, 2019.

\bibitem{li2020online}
Da Li and Timothy Hospedales.
\newblock Online meta-learning for multi-source and semi-supervised domain
  adaptation.
\newblock {\em arXiv preprint arXiv:2004.04398}, 2020.

\bibitem{long2018conditional}
Mingsheng Long, Zhangjie Cao, Jianmin Wang, and Michael~I Jordan.
\newblock Conditional adversarial domain adaptation.
\newblock In {\em Advances in Neural Information Processing Systems}, pages
  1640--1650, 2018.

\bibitem{maaten2008visualizing}
Laurens van~der Maaten and Geoffrey Hinton.
\newblock Visualizing data using t-sne.
\newblock {\em Journal of machine learning research}, 9(Nov):2579--2605, 2008.

\bibitem{park2019relational}
Wonpyo Park, Dongju Kim, Yan Lu, and Minsu Cho.
\newblock Relational knowledge distillation.
\newblock In {\em Proceedings of the IEEE Conference on Computer Vision and
  Pattern Recognition}, pages 3967--3976, 2019.

\bibitem{paszke2017automatic}
Adam Paszke, Sam Gross, Soumith Chintala, Gregory Chanan, Edward Yang, Zachary
  DeVito, Zeming Lin, Alban Desmaison, Luca Antiga, and Adam Lerer.
\newblock Automatic differentiation in pytorch.
\newblock 2017.

\bibitem{peng2019moment}
Xingchao Peng, Qinxun Bai, Xide Xia, Zijun Huang, Kate Saenko, and Bo Wang.
\newblock Moment matching for multi-source domain adaptation.
\newblock In {\em Proceedings of the IEEE International Conference on Computer
  Vision}, pages 1406--1415, 2019.

\bibitem{romero2014fitnets}
Adriana Romero, Nicolas Ballas, Samira~Ebrahimi Kahou, Antoine Chassang, Carlo
  Gatta, and Yoshua Bengio.
\newblock Fitnets: Hints for thin deep nets.
\newblock {\em arXiv preprint arXiv:1412.6550}, 2014.

\bibitem{saito2019semi}
Kuniaki Saito, Donghyun Kim, Stan Sclaroff, Trevor Darrell, and Kate Saenko.
\newblock Semi-supervised domain adaptation via minimax entropy.
\newblock In {\em Proceedings of the IEEE International Conference on Computer
  Vision}, pages 8050--8058, 2019.

\bibitem{saito2017adversarial}
Kuniaki Saito, Yoshitaka Ushiku, Tatsuya Harada, and Kate Saenko.
\newblock Adversarial dropout regularization.
\newblock {\em arXiv preprint arXiv:1711.01575}, 2017.

\bibitem{simonyan2014very}
Karen Simonyan and Andrew Zisserman.
\newblock Very deep convolutional networks for large-scale image recognition.
\newblock {\em arXiv preprint arXiv:1409.1556}, 2014.

\bibitem{ulyanov2016instance}
Dmitry Ulyanov, Andrea Vedaldi, and Victor Lempitsky.
\newblock Instance normalization: The missing ingredient for fast stylization.
\newblock {\em arXiv preprint arXiv:1607.08022}, 2016.

\bibitem{venkateswara2017deep}
Hemanth Venkateswara, Jose Eusebio, Shayok Chakraborty, and Sethuraman
  Panchanathan.
\newblock Deep hashing network for unsupervised domain adaptation.
\newblock In {\em Proceedings of the IEEE Conference on Computer Vision and
  Pattern Recognition}, pages 5018--5027, 2017.

\bibitem{wang2018cosface}
Hao Wang, Yitong Wang, Zheng Zhou, Xing Ji, Dihong Gong, Jingchao Zhou, Zhifeng
  Li, and Wei Liu.
\newblock Cosface: Large margin cosine loss for deep face recognition.
\newblock In {\em Proceedings of the IEEE Conference on Computer Vision and
  Pattern Recognition}, pages 5265--5274, 2018.

\bibitem{xie2020self}
Qizhe Xie, Minh-Thang Luong, Eduard Hovy, and Quoc~V Le.
\newblock Self-training with noisy student improves imagenet classification.
\newblock In {\em Proceedings of the IEEE/CVF Conference on Computer Vision and
  Pattern Recognition}, pages 10687--10698, 2020.

\bibitem{xie2018learning}
Shaoan Xie, Zibin Zheng, Liang Chen, and Chuan Chen.
\newblock Learning semantic representations for unsupervised domain adaptation.
\newblock In {\em International Conference on Machine Learning}, pages
  5423--5432, 2018.

\bibitem{xu2019adversarial}
Minghao Xu, Jian Zhang, Bingbing Ni, Teng Li, Chengjie Wang, Qi Tian, and
  Wenjun Zhang.
\newblock Adversarial domain adaptation with domain mixup.
\newblock {\em arXiv preprint arXiv:1912.01805}, 2019.

\bibitem{yao2015semi}
Ting Yao, Yingwei Pan, Chong-Wah Ngo, Houqiang Li, and Tao Mei.
\newblock Semi-supervised domain adaptation with subspace learning for visual
  recognition.
\newblock In {\em Proceedings of the IEEE conference on computer vision and
  pattern recognition}, pages 2142--2150, 2015.

\bibitem{yuan2020revisiting}
Li Yuan, Francis~EH Tay, Guilin Li, Tao Wang, and Jiashi Feng.
\newblock Revisiting knowledge distillation via label smoothing regularization.
\newblock In {\em Proceedings of the IEEE/CVF Conference on Computer Vision and
  Pattern Recognition}, pages 3903--3911, 2020.

\bibitem{yun2019regularizing}
Sukmin Yun, Jongjin Park, Kimin Lee, and Jinwoo Shin.
\newblock Regularizing class-wise predictions via self-knowledge distillation.
\newblock In {\em The IEEE/CVF Conference on Computer Vision and Pattern
  Recognition (CVPR)}, June 2020.

\bibitem{zagoruyko2016paying}
Sergey Zagoruyko and Nikos Komodakis.
\newblock Paying more attention to attention: Improving the performance of
  convolutional neural networks via attention transfer.
\newblock In {\em ICLR}, 2017.

\bibitem{zhang2019category}
Qiming Zhang, Jing Zhang, Wei Liu, and Dacheng Tao.
\newblock Category anchor-guided unsupervised domain adaptation for semantic
  segmentation.
\newblock In {\em Advances in Neural Information Processing Systems}, pages
  435--445, 2019.

\bibitem{zhou2021domain}
Kaiyang Zhou, Yongxin Yang, Yu Qiao, and Tao Xiang.
\newblock Domain generalization with mixstyle.
\newblock In {\em International Conference on Learning Representations}, 2021.

\end{thebibliography}
}

\newpage


\renewcommand\thesection{\Alph{section}}
\setcounter{section}{0}
\section{Supplemental material}

In this supplementary material, we provide our method's additional details, analyses, and experimental results.

\subsection{Implementation details}
The effectiveness of \ours are shown in different experimental settings in our main paper.
In this section, we provide our experimental details.

Some results of Tables~\ref{tb:domain_all},~\ref{tb:office_home_all}, and~\ref{tb:uda} of our main paper are borrowed from the paper of MME~\cite{saito2019semi}: the results of S+T, DANN, ADR, CDAN, ENT, and MME in Table~\ref{tb:domain_all}, their results with the AlexNet base network in Table~\ref{tb:office_home_all}, and their accuracies of unsupervised domain adaptation in Table~\ref{tb:uda}.

\begin{figure*}[t!]
\centering
\includegraphics[width=\textwidth]{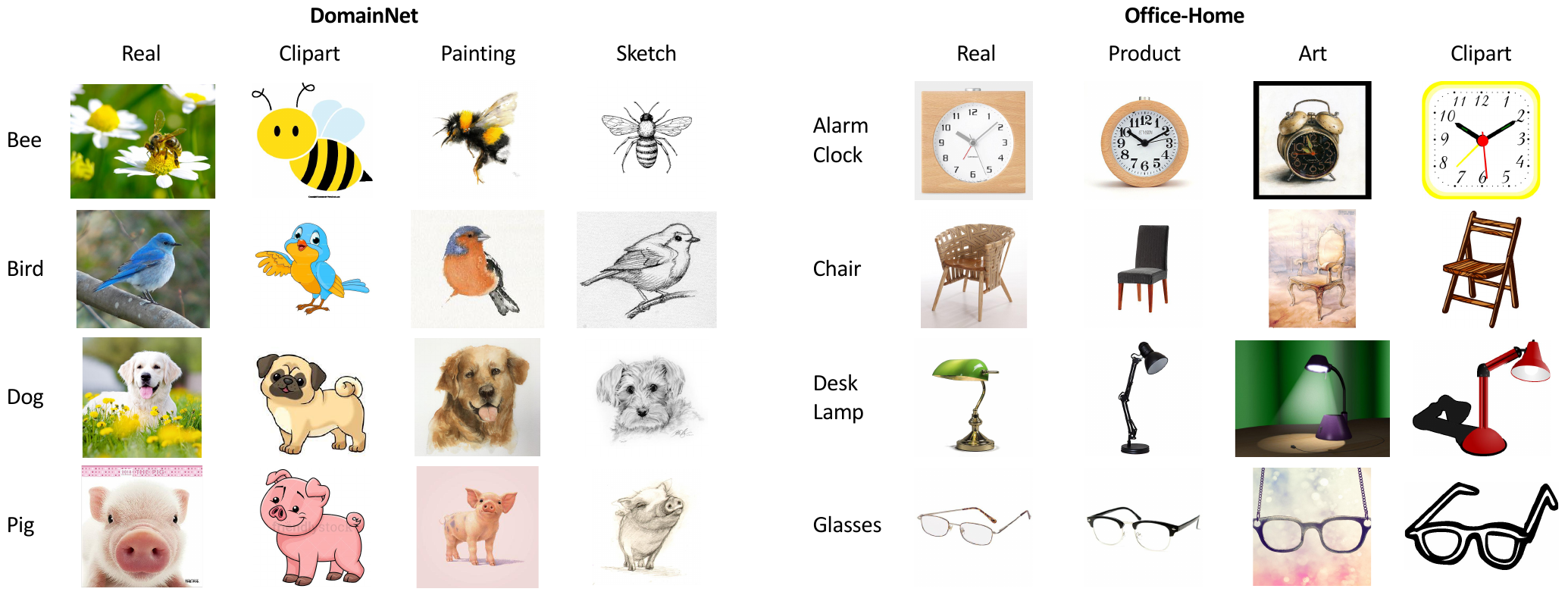}
\caption{DomainNet and Office-Home datasets. 
We visualize four domains of four classes on each dataset.} 
\label{fig:datasets_ex}
\end{figure*}

\paragraph{Datasets.}
In Figure~\ref{fig:datasets_ex}, we visualize the examples of DomainNet and Office-Home datasets. In both datasets, all of four domains are distinct from each other, while Real and Product domains in Office-Home are quite similar.

\paragraph{Baselines.}
For a fair comparison, we reproduce S+T, MME~\cite{saito2019semi}, and APE~\cite{kim2020attract} if the accuracy is not stated in their papers.
To reproduce MME, we follow the official implementation~\footnote{\url{https://github.com/VisionLearningGroup/SSDA_MME}} and set $\lambda$ (from MME) to 0.1.
For APE, we follow the official implementation~\footnote{\url{https://github.com/TKKim93/APE}} and set $\alpha$, $\beta$, and $\gamma$ to 10, 1, and 10, respectively.

\paragraph{Many-shot semi-supervised domain adaptation experiments.} We use ResNet~\cite{he2016deep} for the many-shot experiments on DomainNet dataset in Figure~\ref{fig:many_shot}. The accuracy of the S+T and MME baselines for one-shot and three-shot settings are borrowed from the MME paper.

\paragraph{Unsupervised domain adaptation experiments.} In Table~\ref{tb:uda}, we borrow the accuracy of AlexNet~\cite{krizhevsky2012imagenet} from MME.
We reproduce the accuracy of ResNet for unsupervised domain adaptation experiments under controlled settings.

\paragraph{Ablation studies.} 
We conduct our experiments on DomainNet using ResNet34 for ablation experiments shown in Tables~\ref{table:student} and \ref{tb:analysis_ablation}.
Table~\ref{tb:analysis_ablation} indicates the full ablation study of proposed components.
In Figure~\ref{fig:ablation_components}, we use AlexNet for our base network and DomainNet for our dataset.

\paragraph{Inter-domain and intra-domain discrepancy histograms.}
We use ResNet for Office-Home~\cite{venkateswara2017deep} on one-shot setting to plot inter-domain and intra-domain discrepancy histograms shown in Figure~\ref{fig:hist_all}.
We choose the Clipart domain for the source domain and the Product domain for the target domain.
For Figures \ref{fig:inter} and \ref{fig:intra}, we plot histograms of cosine similarity after the pre-training stage, specifically from 10,000$^{\text{th}}$ iteration.
We plot the histogram every 3,000 iterations until the model converges.
For APE, we plot the converged model for a comparison.

\begin{figure}
\begin{center}
\includegraphics[width=0.80\linewidth]{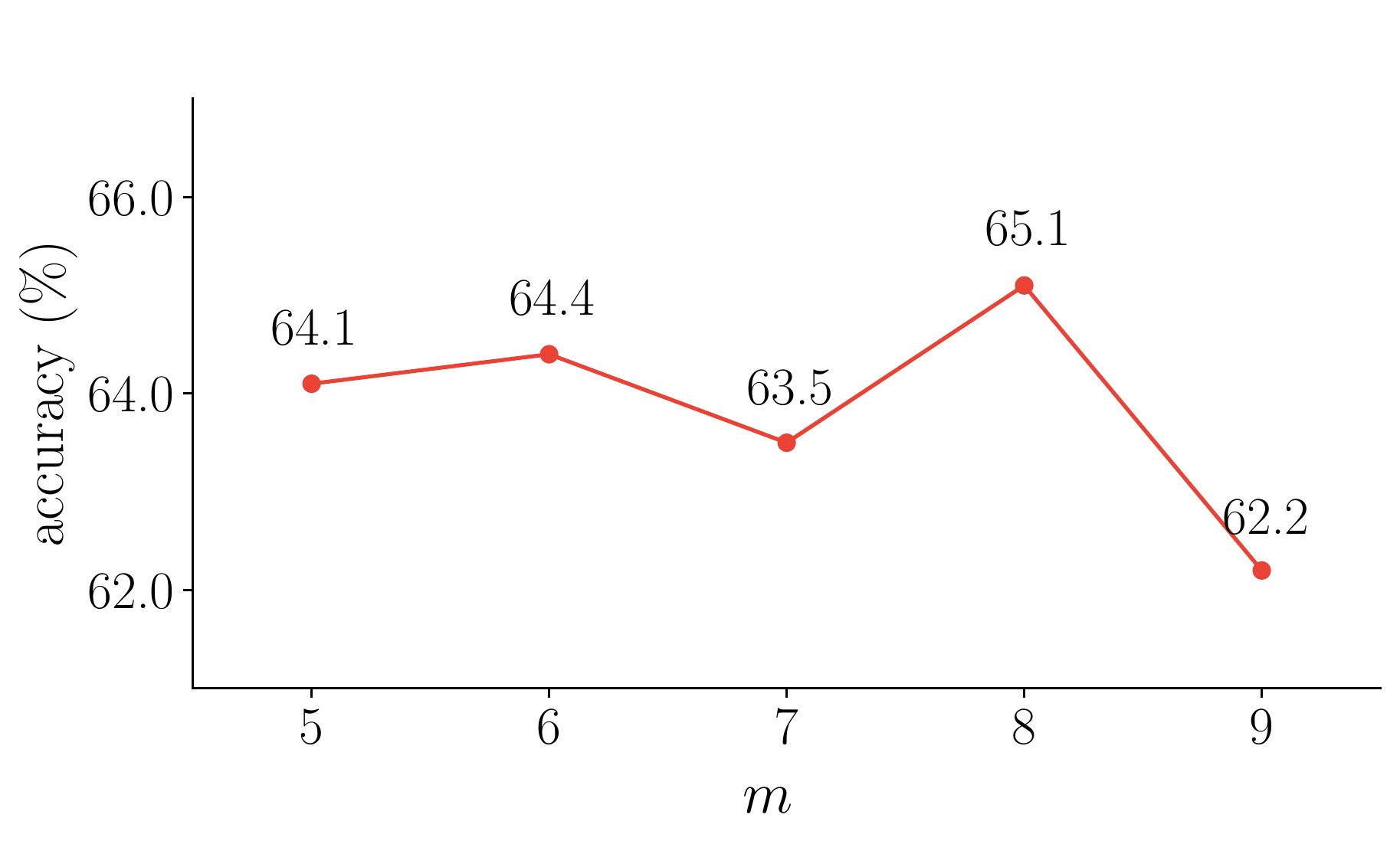}
\end{center}
\vspace{-6mm}
\caption{Accuracy (\%) in Real to Sketch one-shot scenario on the DomainNet using ResNet with various $m$. \label{fig:hyper}}
\end{figure}

\paragraph{The balancing hyper-parameter $\lambda$.}
\label{sec:hyper}
We use $\lambda$ to balance $\mathcal{L}_{\text{pair}}$ in the overall loss and make up for the incompleteness of pseudo-labels.
We set the hyper-parameter $\lambda$ using a ramp-up function like in \cite{ganin2016domain}:
\begin{eqnarray}
\lambda = \frac{2}{1+e^{-mt}} - 1,
\end{eqnarray}
where $t \in [0, 1]$ increases over iterations.
The increasing $t$ makes $\lambda$ increases so that $\mathcal{L}_{\text{pair}}$ influences more on the learning process.
This incremental weighting technique is adequate since pseudo-labels are likely to be incorrect at the beginning of the sample-to-sample training stage.

To find a proper ramp-up function, we vary $m$ to examine the effects of weighting $\mathcal{L}_\text{pair}$.
Changing $m$ controls the slope of the ramp-up function.
We choose Real to Sketch one-shot domain scenario, and select $m$ at the best validation accuracy.
In DomainNet and Office-Home experiment, we set $m$ to 8 on both AlexNet and ResNet.
Figure~\ref{fig:hyper} shows the accuracy of our model when varying $m$.

\begin{figure}[t!]
  \begin{subfigure}{0.233\textwidth}
    \includegraphics[width=\linewidth]{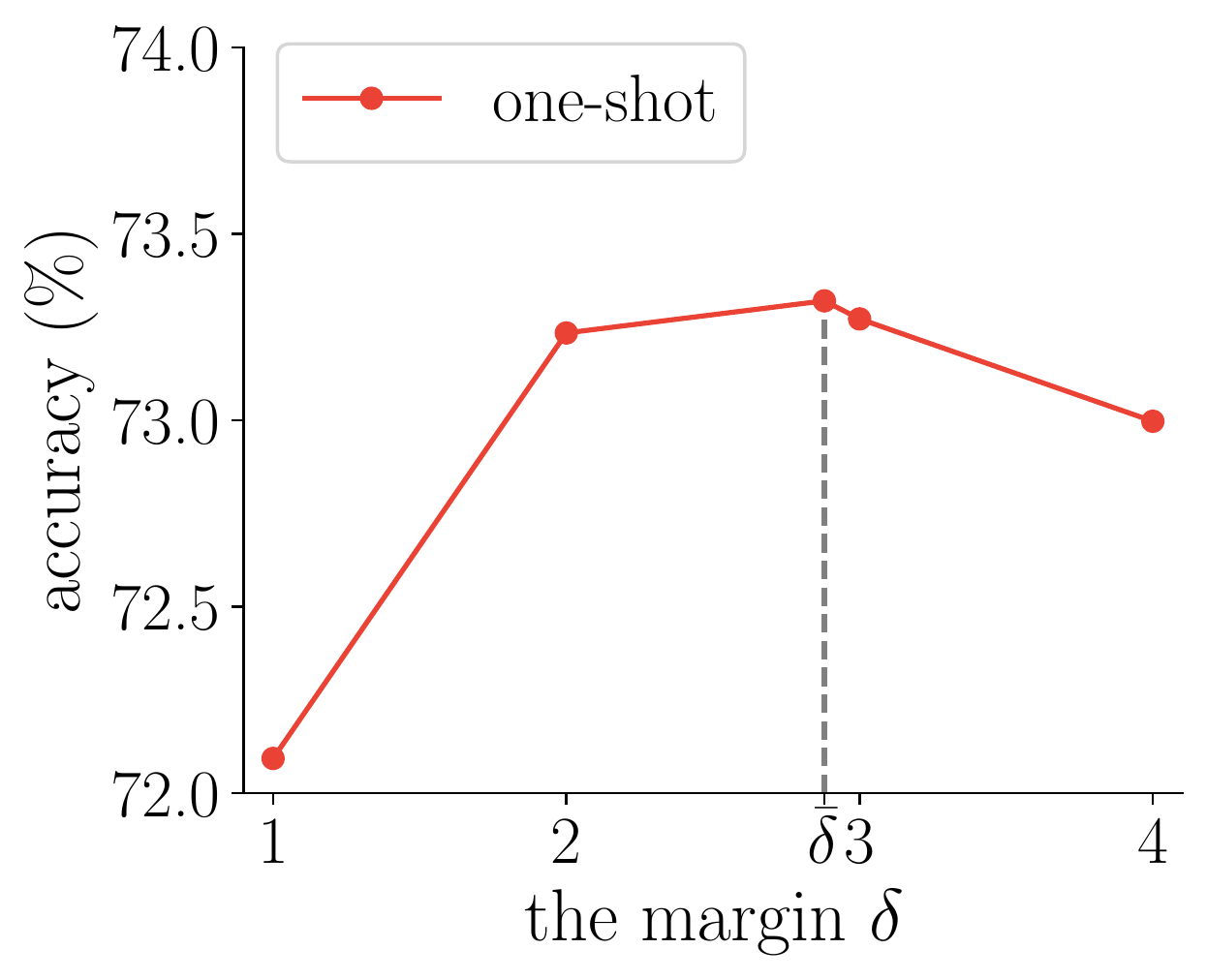}
    \vspace{-6mm}
    \caption{One-shot setting. $\bar{\delta}$=2.88.  \label{fig:delta_1shot}}
  \end{subfigure}
  \hspace*{\fill}   
  \begin{subfigure}{0.233\textwidth}
    \includegraphics[width=\linewidth]{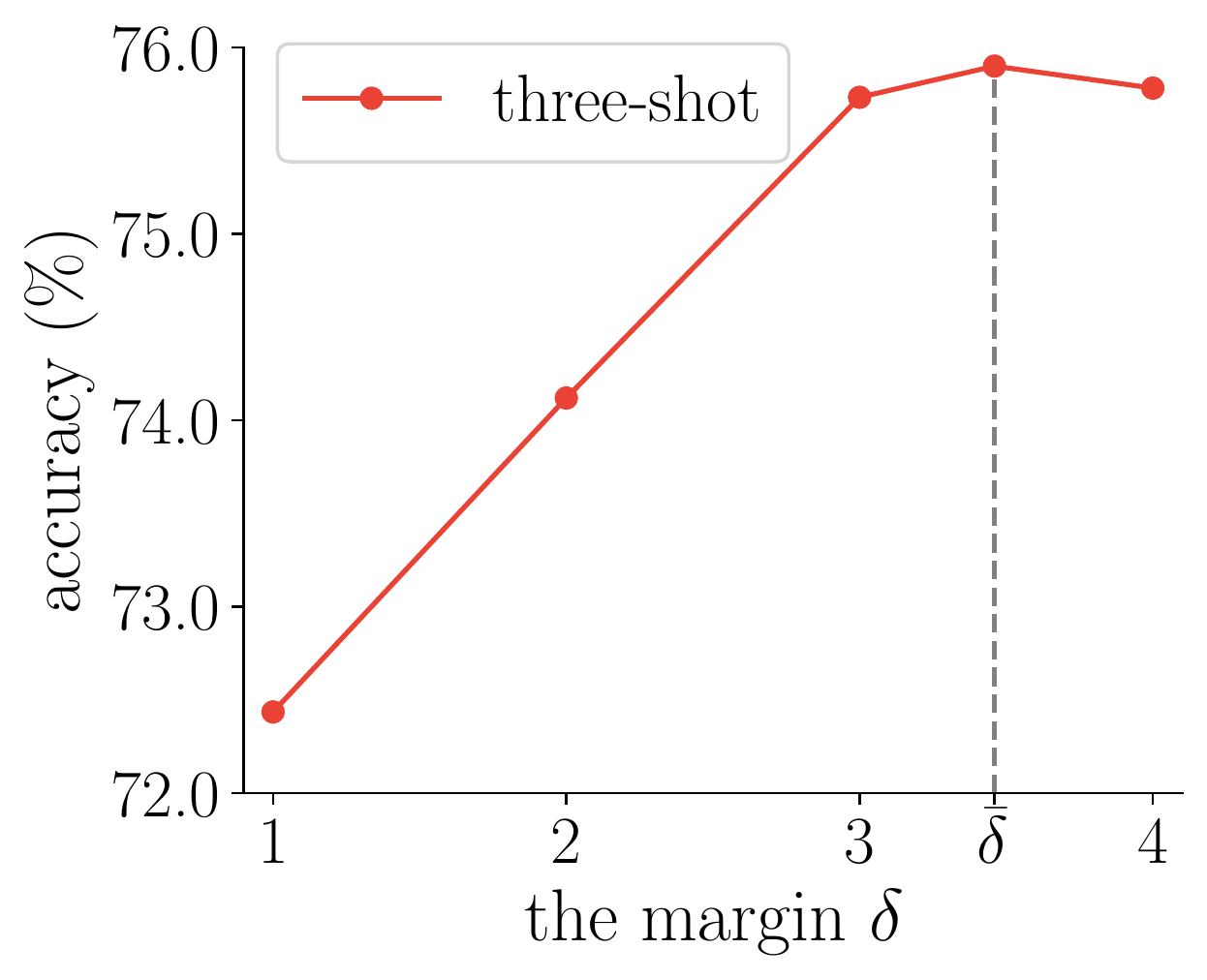}
    \vspace{-6mm}
    \caption{Three-shot setting. $\bar{\delta}$=3.46. \label{fig:delta_3shot}}
  \end{subfigure}
\caption{
Accuracy (\%) on Real to Clipart (DomainNet) scenario using ResNet34 with various $\delta$.
\label{fig:margin_study}}
\end{figure}

\paragraph{The class logit margin $\delta$.}
In~\Eqref{eq:reliability_evaluation}, the margin $\delta$ is used to filter out unreliable target samples.
Here, $\delta$ determines a trade-off between a number of pseudo-labels used and how reliable the pseudo-labels are.
A small $\delta$ makes the pseudo-labels of the student set inaccurate. 
On the contrary, a large $\delta$ makes RSS filter many unlabeled target samples so that few student samples are used for training.
Therefore, proper $\delta$ is critical for a student-set to contain precise and various student samples.

We investigate whether the average margin $\bar{\delta}$ of all unlabeled target samples is appropriate for $\delta$ or not.
Figure~\ref{fig:margin_study} shows the experiment, which is conducted on DomainNet Real to Clipart scenario using ResNet34.
We compare the result of the average margin $\bar{\delta}$ to those of the margin $\delta$ from 1 to 4.
$\bar{\delta}$ is initially calculated from the pre-trained model and is fixed afterward. 
In Figure~\ref{fig:margin_study}, the model calculates $\bar{\delta}$ as 2.88 and 3.46 in one-shot and three-shot setting, respectively.
The model shows the best accuracy when the student-set is generated using $\bar{\delta}$.
This result describes that $\bar{\delta}$ is an appropriate margin for RSS to make student-set abundant and precise.
\\

\begin{table}[t]
\begin{center}
\scalebox{0.9}{
\begin{tabular}{l|ccccc}
\toprule[1pt] 
 \multirow{2}{*}{Method}  &\multicolumn{5}{c}{$\delta$}\\ 
 &1 &2 &3 &$\bar{\delta}$ &4 \\ \hline
 \multirow{1}{*}{\ours ($\alpha = 0.7$)} & 72.6 & 72.9 & 73.1 & 73.5 & 73.4 \\
 \multirow{1}{*}{\ours ($\alpha = 0.8$)} & 72.5 & 74.4 & 75.0 & \bf{75.9} & 75.7 \\
\multirow{1}{*}{\ours ($\alpha = 0.95$)} & 72.4 & 74.1 & 75.7 & \bf{75.9} & 75.8 \\
\multirow{1}{*}{\ours w/o $\alpha$} & 72.2 & 72.9 & 75.2 & 75.6 & 75.5 \\
\bottomrule[1pt]
 \end{tabular}}
\end{center}
\vspace{-6mm}
\caption{Accuracy (\%) on Real to Clipart (DomainNet) three-shot scenario using ResNet34 by varying $\delta$ and $\alpha$. \label{tb:optimal}}
\vspace{-2mm}
\end{table}

\paragraph{Performance varying both $\delta$ and $\alpha$.}
Unlike \cite{zhang2019category}, we preset the margin $\delta$ by averaging all unlabeled target's margin so that we can obtain target adaptive $\delta$.
This is because \ours deals with various target domains different from \cite{zhang2019category}, which considers only Cityscapes~\cite{cordts2016cityscapes} as a target dataset.
Also, the threshold $\alpha$ is designed to avoid the situation that CAG~\cite{zhang2019category} might exclude the sample with high confidence because of its low margin.
To search the best values of $\delta$ and $\alpha$, we jointly vary the values of them.
The details are the same as the setting in Figure \ref{fig:delta_3shot}.
The results are shown in Table~\ref{tb:optimal}, where \ours ($\alpha=0.95$, $\delta=\bar{\delta}$) performs the best.
It also shows that \ours is not sensitive to the margin parameter $\delta$ and the threshold $\alpha$.

\begin{figure}
\centering
\includegraphics[width=0.28\textwidth]{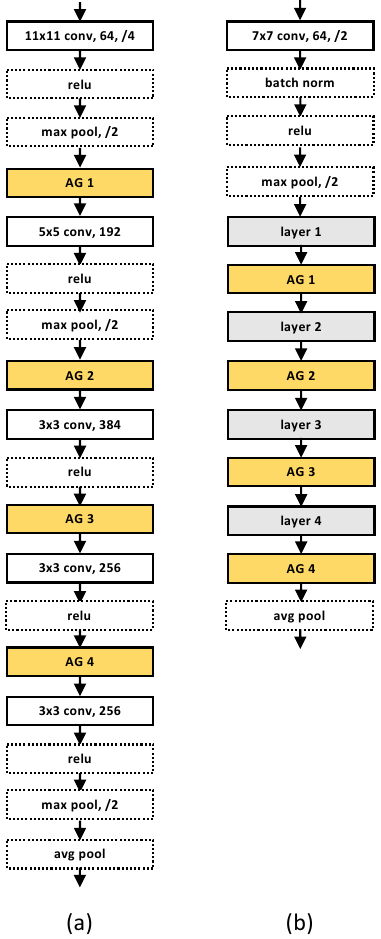}
\caption{
The locations to apply AG operation.
The yellow blocks represent the place to apply AG.
(a) The feature extractor of AlexNet.
(b) The feature extractor of ResNet34.
} 
\label{fig:location}
\end{figure}

\paragraph{The locations to apply AG.}
To generate assistant features, AG operation is applied to different layers.
In AlexNet, the operation can be placed at any yellow blocks shown in Figure~\ref{fig:location} (a).
In ResNet34, the operation can be located at the end of every four residual blocks shown in Figure~\ref{fig:location} (b).
We conduct experiments on various combinations of the location in Table~\ref{tb:location}.
We conjecture that the optimal combination of the operations is different between ResNet and AlexNet. 
As the overall accuracy of ResNet is higher than that of AlexNet, 
the pairs from ResNet are more reliable and thus the semantic meanings from teachers are more effective for students in ResNet than AlexNet.
We select all locations for ResNet in DomainNet and Office-Home.
For AlexNet, we select AG 1 and AG 1,2 in DomainNet and Office-Home, respectively.

\begin{table}[t]
\begin{center}
\scalebox{0.86}{
\begin{tabular}{l|cccc}
\toprule[1pt] 
 \multirow{2}{*}{Network}  &\multicolumn{4}{c}{AG}\\ 
 &1 &1,2 &1,2,3 &1,2,3,4 \\ \hline
\multirow{1}{*}{AlexNet} & \bf{39.9}  & 38.5 & 31.8 & 32.6 \\
\multirow{1}{*}{ResNet34} & 60.5  & 61.2 & 62.5 & \bf{65.1} \\
\bottomrule[1pt]
 \end{tabular}}
\end{center}
\vspace{-4mm}
\caption{Accuracy (\%) in two networks on the DomainNet Real to Sketch one-shot scenario with various location combinations of AG operation. \label{tb:location}}
\end{table}

\begin{table}[ht]
\vspace{2mm}
\begin{center}
\scalebox{0.8}{
\begin{tabular}{l|l|cc}
\toprule[1pt]
Net& Method       & DomainNet & Office-Home \\ \hline
\multirow{4}{*}{AlexNet} 
& S+T & 40.0 & 44.1 \\
& MixStyle & 39.3 & 43.5  \\\cline{2-4}
& \ours ($\epsilon=1$) & 46.7 & 46.8 \\
&\ccol\ours (ours) & \ccol 48.7 & \ccol 49.5 \\\hline

\multirow{4}{*}{ResNet34}
& S+T & 56.9 & 62.3 \\
& MixStyle & 66.6 & 60.2  \\\cline{2-4}
& \ours ($\epsilon=1$) & 69.1 & 69.8  \\
&\ccol\ours (ours) & \ccol 69.9 & \ccol 70.3 \\
\bottomrule[1pt]
\end{tabular}}
\end{center}
\vspace{-4mm}
\caption{Average classification accuracy (\%) on the DomainNet and Office-Home datasets for one-shot on all domain scenarios that we cover. \label{tb:style}}
\end{table}

\begin{table}[t]
\begin{center}
\scalebox{0.9}{
\begin{tabular}{l|cc}
\toprule[1pt] 
 \multirow{1}{*}{Method}  &1-shot & 3-shot\\ \hline
\multirow{1}{*}{CDAN} & 62.9$\pm{1.5}$ & 65.3$\pm{0.1}$ \\
\multirow{1}{*}{ENT} & 59.5$\pm{1.5}$ & 63.6$\pm{1.3}$ \\
\multirow{1}{*}{MME} & 64.3$\pm{0.8}$ & 66.8$\pm{0.4}$  \\
\multirow{1}{*}{APE} & 65.2$\pm{0.9}$ & 67.3$\pm{0.9}$  \\
\multirow{1}{*}{\ccol \ours} & \ccol 67.7$\pm{0.6}$ & \ccol 69.7$\pm{0.7}$  \\
\bottomrule[1pt]
 \end{tabular}}
\end{center}
\vspace{-4mm}
\caption{Classification accuracy (\%) and standard deviation (\%) on the Sketch to Painting scenario in the DomainNet averaged over three runs.\label{tb:multiple_run} }
\end{table}

\begin{table*}[ht]
\vspace{-2mm}
\begin{center}
\scalebox{0.93}{
\begin{tabular}{l|ccc|ccccccc|c}
\toprule[1pt]
Method & $\mathcal{L}_\text{unl}$ & $\mathcal{L}_\text{pair}$ & RSS & R to C & R to P & P to C & C to S & S to P & R to S & P to R & MEAN\\ \hline

 DANN   & & &      & 58.2 & 61.4 & 56.3 & 52.8 & 57.4 & 52.2 & 70.3 & 58.4 \\
 MME    & & &      & 70.0 & 67.7 & 69.0 & 56.3 & 64.8 & 61.0 & 76.1 & 66.4 \\
 APE    & & &     & 70.4 & 70.8 & 72.9 & 56.7 & 64.5 & 63.0 & 76.6 & 67.6 \\\cline{1-12}
 \multirow{7}{*}{\ours} 
 &\xmark & \xmark & \xmark &56.8  & 60.5 & 55.4 & 51.7 & 55.5 & 47.5 & 72.0 & 57.1\\\cline{2-12}
 &\cmark & \xmark & \xmark & 68.7 & 65.6 & 68.8 & 59.2 & 64.1 & 61.6 & 78.4 & 66.6 \\
 &\cmark & \xmark &\cmark & 71.6 & 69.1 & 70.7 & 58.7 & 65.4 & 62.0 & 79.6 & 68.2 \\\cline{2-12}
 & \xmark &\cmark& \xmark & 67.4 & 65.0 & 67.1 & 61.2 & 64.9 & 62.7 & 77.5 & 66.5 \\
 & \xmark &\cmark&\cmark & 73.1 & 67.1 & 70.6 & 57.7 & 65.8 & 62.4 & 73.6 & 67.2 \\ \cline{2-12}
 &\cmark &\cmark& \xmark & 69.4 & 65.7 & 69.7 & 61.3 & 65.5 & 61.7 & 78.6 & 67.4 \\
 &\ccol \cmark &\ccol \cmark&\ccol \cmark & \ccol 73.3  & \ccol 68.9 & \ccol 73.4 & \ccol 60.8 & \ccol 68.2 & \ccol 65.1 & \ccol 79.5 & \ccol 69.9  \\
\bottomrule[1pt]
\end{tabular}}
\end{center}
\caption{Comprehensive ablation study of \ours on DomainNet dataset (\%) for one-shot setting.\label{tb:analysis_ablation}
}
\end{table*}

\begin{table}[t!]
\begin{subfigure}{\linewidth}
\centering
\scalebox{0.84}{
\begin{tabular}{l|cccccc}
\toprule[1pt] 
Method  &0-shot &1-shot & 3-shot & 5-shot & 10-shot & 20-shot\\ \midrule
S+T  & 54.5 & 55.6 & 60.0 & 64.6 & 67.6 & 71.5 \\
MME  & 67.6 & 70.0 & 72.2 & 74.8 & 76.8 & 77.9 \\
APE & 65.4 & 67.0 & 72.2 & 72.8 & 76.9 & 77.0 \\ 
\ccol \ours& \ccol \textbf{72.7} & \ccol \textbf{73.3} & \ccol \textbf{75.9} & \ccol \textbf{77.5} & \ccol \textbf{78.0} & \ccol \textbf{79.1} \\
\bottomrule[1pt]
\end{tabular}}
\caption{Real to Clipart. \label{tb:many_shot_numbers_r_to_c}}
\end{subfigure}
\begin{subfigure}{\linewidth}
\vspace{2mm}
\centering
\scalebox{0.84}{
\begin{tabular}{l|cccccc}
\toprule[1pt] 
Method  &0-shot &1-shot & 3-shot & 5-shot & 10-shot & 20-shot\\ \midrule
S+T  & 55.9 & 56.8 & 59.4 & 64.4 & 68.4 & 71.1 \\
MME  & 67.1 & 69.0 & 71.7 & 73.0 & 76.4 & 78.0 \\
APE & 63.8 & 67.7 & 71.3 & 72.6 & 76.6 & 78.1 \\ 
\ccol \ours & \ccol \textbf{66.5} & \ccol \textbf{73.4} & \ccol \textbf{75.1} & \ccol \textbf{76.9} & \ccol \textbf{77.2} & \ccol \textbf{79.6} \\
\bottomrule[1pt]
\end{tabular}}
\caption{Painting to Clipart. \label{tb:many_shot_numbers_p_to_c}}
\end{subfigure}
\caption{Classification accuracy (\%) on DomainNet with a varying number of target labels.
(a) corresponds to Figure~\ref{fig:many_r_to_c2}, and (b) corresponds to Figure~\ref{fig:many_p_to_c2}.
\label{tb:many_shot_numbers_all}}
\end{table}

\subsection{Additional experimental results}

\paragraph{Comparison with MixStyle~\cite{zhou2021domain}.}
The main difference between \ours and MixStyle is that we introduce the assistant (intermediate style feature) as a guidance for the student. 
The assistant is designed to transfer its knowledge to the student using knowledge distillation; for this reason, we do not back-propagate gradients
through the path of assistant features (see the second dotted branch in Figure \ref{fig:architecture}). 
This strategy has not been explored before. 
MixStyle, which is introduced for domain generalization, directly trains the model with stylized features; the features’ predictions and
given labels are used for calculating the cross-entropy loss,
and the gradients are back-propagated through the features.
This scheme is not adequate for SSDA for the reason that the goal of SSDA is to adapt the learner to the target domain. 
For comparison, we conduct experiments where we directly apply the
scheme of \cite{zhou2021domain} to SSDA; we only change the $\mathcal{L}_\text{pair}$ loss to
the cross-entropy loss between assistants’ predictions and pseudo-labels.
We also searched the best hyper-parameters for this model as we did for \ours. 
We set $m$ to 9 for all experiments.
For ResNet, we select AG 1,2,3 for DomainNet and AG 1,2,3,4 for Office-Home.
For AlexNet, we select AG 1 for DomainNet and AG 1,2 for Office-Home. 
In Table~\ref{tb:style}, it is obvious that MixStyle is not effective except for the experiment of ResNet in DomainNet.
The model even shows low accuracy than the simple baseline S+T in several experiments.

\paragraph{The effect of intermediate styles.}
In Table~\ref{tb:style}, we examine the effectiveness of transferring an intermediate style rather than a teacher's individual style.
In \Eqref{eq:style-fusion}, by controlling the value of $\epsilon$, we can manipulate the style of the assistant feature.
As the value of $\epsilon$ is close to 1, the style of the assistant approaches to the teacher's one.
\ours($\epsilon=1$) is the experiment that the assistant feature follows only the style of the teacher.
When we compare \ours($\epsilon=1$) and \ours, the results show that the intermediate styles are more effective than the teacher's style to reduce the domain discrepancy.

\paragraph{Multiple runs.}
For a fair comparison, we report the average accuracy and its standard deviation of three independent runs in Table~\ref{tb:multiple_run}.
Our method less deviates than most of previous methods do, showing that our method is adequately stable and effective.

\paragraph{Many-shot experiments.} In Table~\ref{tb:many_shot_numbers_all}, we attach exact values plotted in Figure~\ref{fig:many_shot} of the main paper.

\begin{figure*}[ht]
  \begin{subfigure}{\textwidth}
\centering
    \includegraphics[width=0.85\linewidth]{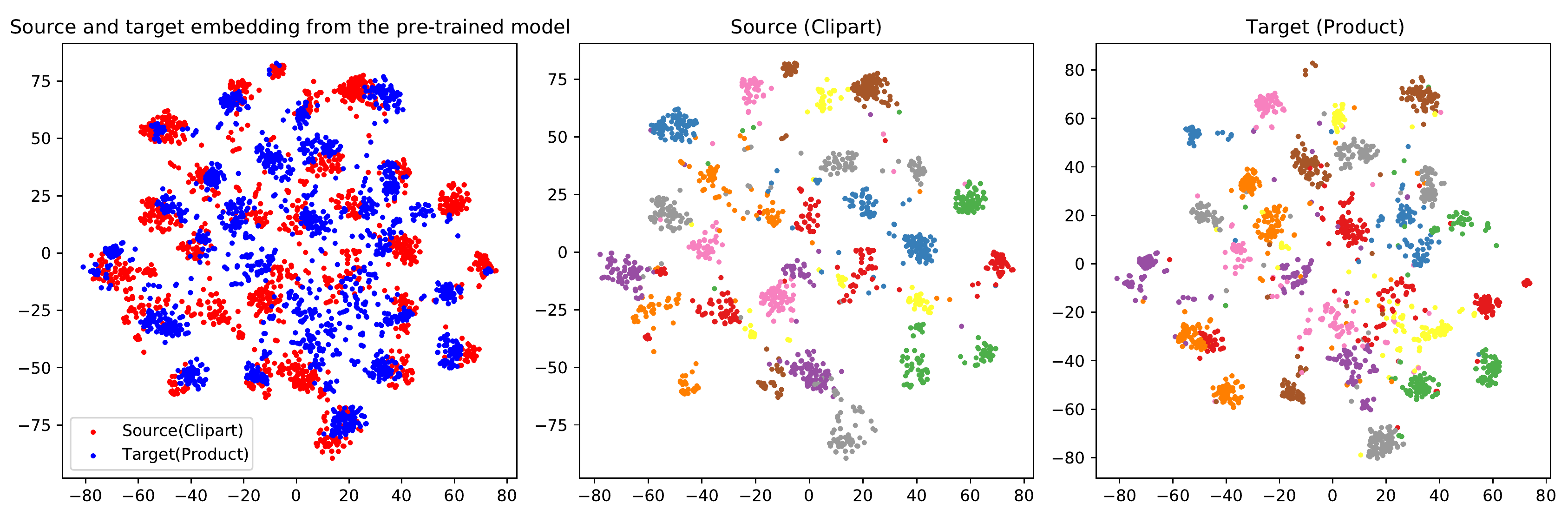}
    \vspace{-2mm}
    \caption{$t$-SNE visualization after pre-training stage on Clipart $\rightarrow$ Product.  \label{fig:tsne_pre1}}
  \end{subfigure}
  \hspace*{\fill} 
  \vfill
  \begin{subfigure}{\textwidth}
\centering 
    \includegraphics[width=0.85\linewidth]{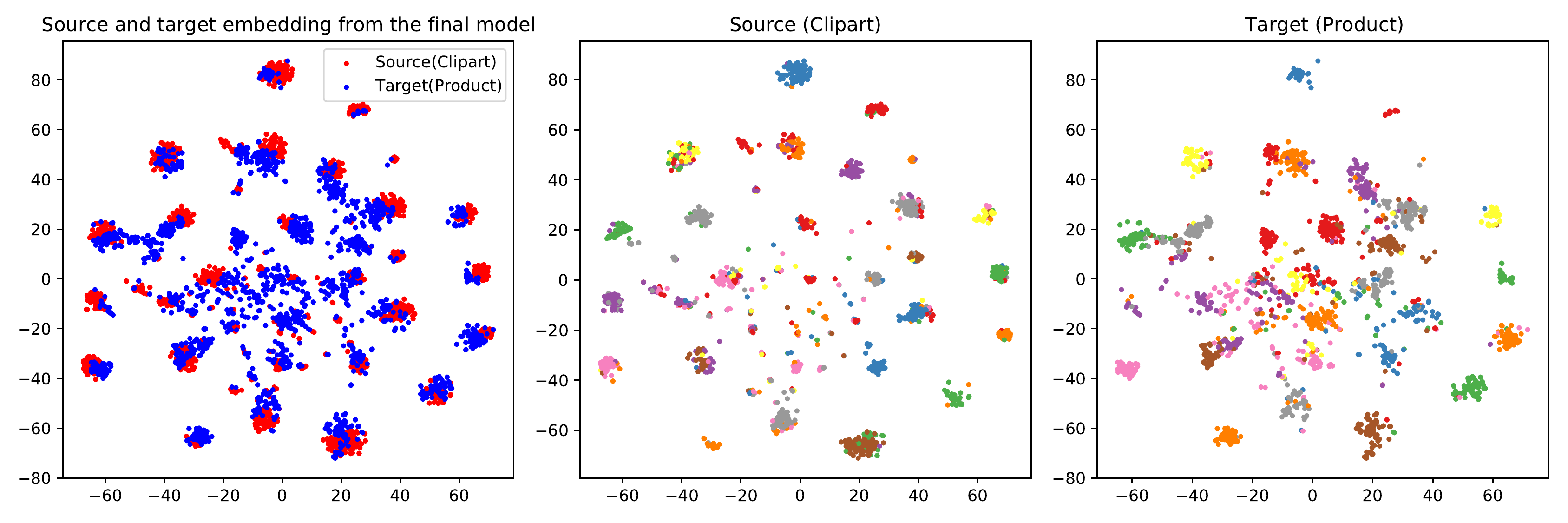}
    \vspace{-2mm}
    \caption{$t$-SNE visualization at the final model on Clipart $\rightarrow$ Product. \label{fig:tsne_final1}}
  \end{subfigure}
  \hspace*{\fill}   
  \vfill
  \begin{subfigure}{\textwidth}
\centering 
    \includegraphics[width=0.85\linewidth]{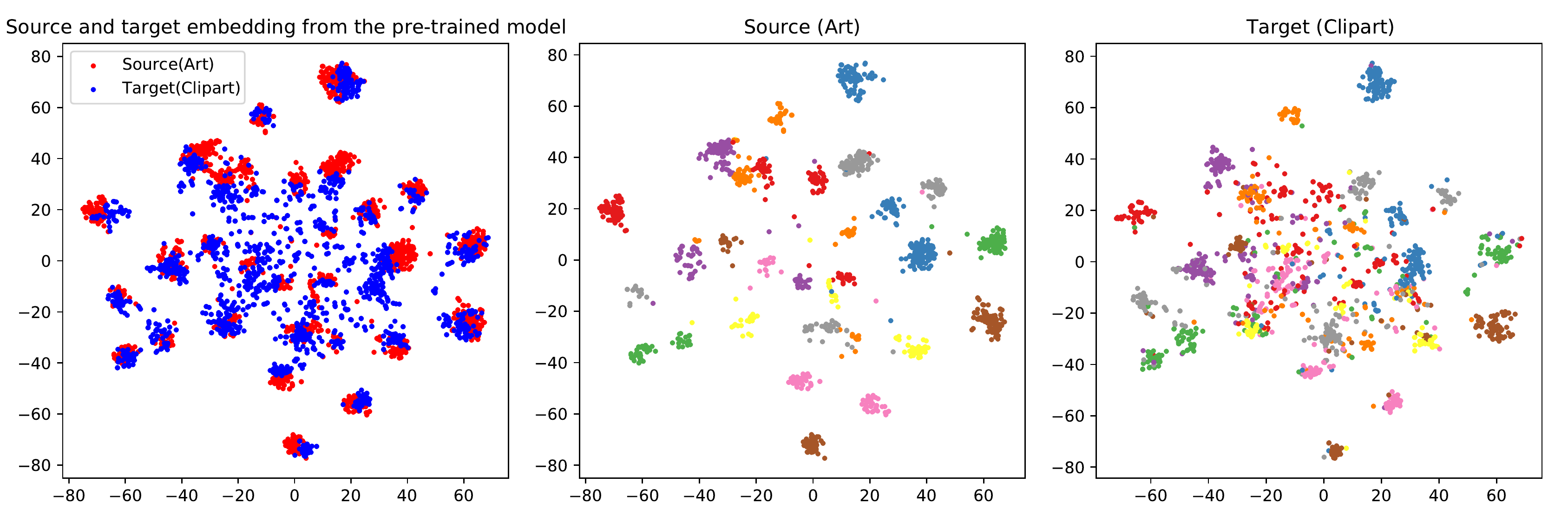}
    \vspace{-2mm}
    \caption{$t$-SNE visualization after pre-training stage on Art $\rightarrow$ Clipart.  \label{fig:tsne_pre2}}
  \end{subfigure}
  \hspace*{\fill} 
  \vfill
  \begin{subfigure}{\textwidth}
\centering 
    \includegraphics[width=0.85\linewidth]{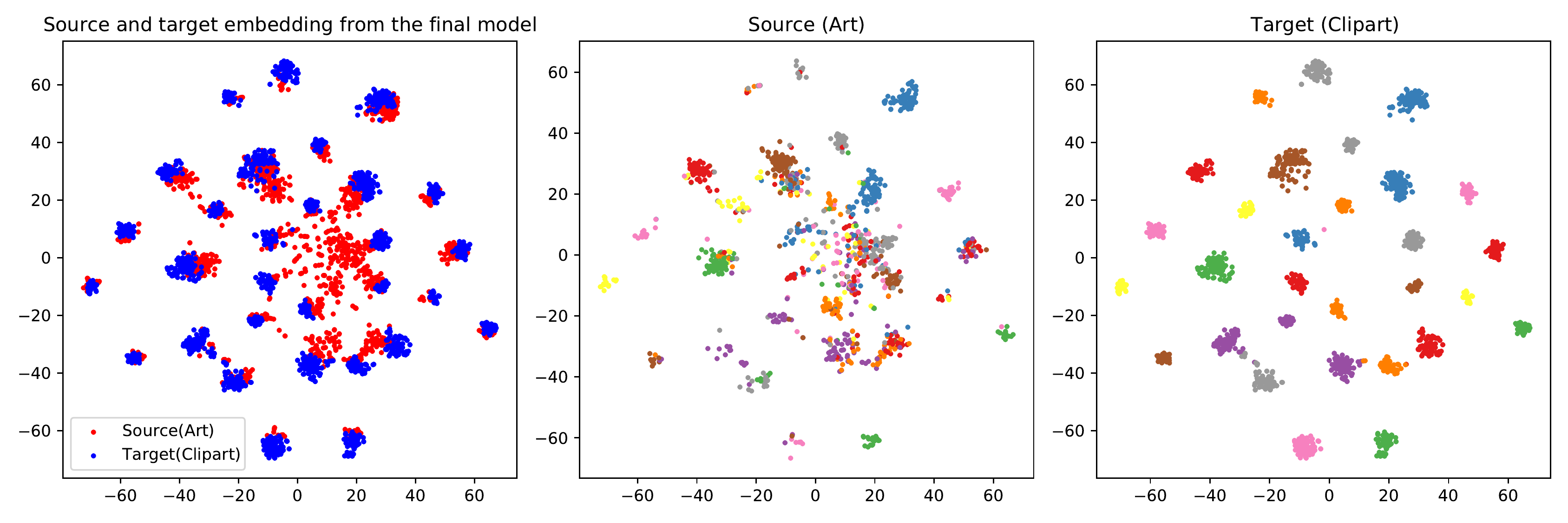}
    \vspace{-2mm}
    \caption{$t$-SNE visualization at the final model on Art $\rightarrow$ Clipart.\label{fig:tsne_final2}}
  \end{subfigure}
  \hspace*{\fill} 
\caption{ $t$-SNE visualization on the Office-Home. Left column: source and target embedding spaces. Middle column: source embedding spaces. Right column: target embedding spaces.
\label{fig:tsne_all_sup}}
\end{figure*}

\paragraph{Extra $t$-SNE visualization}
Figure~\ref{fig:tsne_all_sup} visualizes how \ours embeds instances from two domains over iterations.
The embeddings are obtained using ResNet34 from examples of Office-Home dataset in the one-shot setting, and we visualize the first 30 classes for simplicity.
We adopt $t$-SNE~\cite{maaten2008visualizing} with the perplexity of 30.0 and 1000 iterations.
We observe that the sample-to-sample self-distillation stage clearly enhances the embedding quality from the pre-training stage.
Two main points of the results are: (1) target samples gradually align with source samples over iterations. (2) samples from the same class pull each other over iterations.

\paragraph{Comprehensive ablation study.} 
We evaluate different combinations of the proposed component in Table~\ref{tb:analysis_ablation} and compare them with previous work of~\cite{saito2019semi, ganin2016domain, kim2020attract}.
The performance consistently increases as more components are used, indicating that each proposed component is effective for SSDA.
Note that our method with all the components sets a new state of the art, outperforming APE~\cite{kim2020attract}.  

\end{document}